\documentclass[11pt]{article}

\usepackage[final]{acl}

\usepackage{times}
\usepackage{latexsym}

\usepackage[T1]{fontenc}

\usepackage[utf8]{inputenc}

\usepackage{microtype}

\usepackage{inconsolata}
\usepackage{subcaption}
\usepackage{graphicx}
\usepackage[dvipsnames]{xcolor}
\usepackage{booktabs}       
\usepackage{amsfonts}       
\usepackage{nicefrac}       
\usepackage{microtype}      
\usepackage{pifont}
\usepackage[table]{xcolor}
\usepackage{booktabs}
\usepackage{multirow}
\usepackage{array}
\usepackage{pifont}   
\usepackage{enumitem} 
\usepackage{amsmath}
\usepackage{amssymb}
\usepackage{multirow}
\usepackage[table]{xcolor}
\usepackage[most]{tcolorbox}
\newcommand{\red}[1]{#1}
\definecolor{mygray}{gray}{.9}
\definecolor{MyBlue}{RGB}{0,172,238}
\definecolor{MyPink}{RGB}{237,0,139}
\definecolor{headergray}{RGB}{227,226,233}
\definecolor{rowgray}{RGB}{242,242,242}
\hypersetup{
	colorlinks=true,
	linkcolor=red,
	citecolor=MyBlue,
	urlcolor=MyPink
}

\usepackage{url}            
\usepackage{booktabs}       
\usepackage{amsfonts}       
\usepackage{nicefrac}       
\usepackage{microtype}      
\usepackage{xcolor}         

\definecolor{findingboxbg}{RGB}{242,242,242}
\definecolor{findingboxframe}{RGB}{70,70,70}

\newtcolorbox{findingbox}{
	colback=findingboxbg,
	colframe=findingboxframe,
	boxrule=1.1pt,
	arc=5pt,
	left=7pt,
	right=7pt,
	top=6pt,
	bottom=6pt,
	boxsep=0pt,
	before skip=6pt,
	after skip=8pt,
	enhanced
}

\definecolor{headergray}{RGB}{230,230,236}
\definecolor{rowgray}{RGB}{242,242,242}
\definecolor{upred}{RGB}{200,60,60}
\definecolor{downgreen}{RGB}{60,140,80}

\definecolor{boxpurple}{RGB}{126,123,255}
\definecolor{boxbg}{RGB}{235,235,255}

\newtcolorbox{promptbox}[1]{
	enhanced,
	breakable,
	colback=boxbg,
	colframe=boxpurple,
	coltitle=white,
	colbacktitle=boxpurple,
	title={#1},
	fonttitle=\bfseries\large,
	fontupper=\large,
	boxrule=1.0pt,
	arc=2.5mm,
	left=6mm,
	right=6mm,
	top=4mm,
	bottom=4mm,
	toptitle=2mm,
	bottomtitle=2mm,
	titlerule=0pt,
	before skip=8pt,
	after skip=10pt
}
\usepackage{cuted}
\usepackage[checkfloat]{flushend}

\tcbuselibrary{listings,breakable,skins}

\newtcblisting{compactprompt}[1]{
	enhanced,
	breakable,
	listing only,
	colback=gray!4,
	colframe=black!45,
	coltitle=black,
	colbacktitle=gray!18,
	title={#1},
	fonttitle=\bfseries\small,
	boxrule=0.4pt,
	arc=1.2mm,
	left=4pt,
	right=4pt,
	top=4pt,
	bottom=4pt,
	before skip=5pt,
	after skip=6pt,
	listing options={
		basicstyle=\scriptsize\ttfamily,
		breaklines=true,
		breakatwhitespace=false,
		columns=fullflexible,
		keepspaces=true,
		showstringspaces=false,
		upquote=true,
		tabsize=2
	}
}

\makeatletter
\newcommand{\restoreflushbottomglobally}{%
	\global\let\@textbottom\relax
	\global\let\@texttop\relax
}
\makeatother

\title{Verification Trap: Understanding Test-Time Selection Failures under False Premises in Code Generation}

\author{Feng He\textsuperscript{1}, Hejia Wang\textsuperscript{2}, Linghao Meng\textsuperscript{3}, Ming Gao\textsuperscript{1}, Qiankun Li\textsuperscript{4*} \\
  {\normalfont\textsuperscript{1}University of Science and Technology of China, China} \\
  {\normalfont\textsuperscript{2}Beijing University of Posts and Telecommunications, China} \\
  {\normalfont\textsuperscript{3}National University of Singapore, Singapore} \\
  {\normalfont\textsuperscript{4}Nanyang Technological University, Singapore} \\
  {\normalfont\texttt{hefengcs@gmail.com, cs-qiankun.li@ntu.edu.sg}}}

\begin{document}
\flushbottom
\maketitle
\begingroup
\renewcommand{\thefootnote}{}
\footnotetext{\textsuperscript{*}Corresponding author.}
\endgroup
\begin{abstract}
Test-time compute has become a central way to improve code generation: systems sample multiple candidate programs and use verifier-visible evidence to select the final output. This paradigm implicitly assumes that the verifier provides a corrective signal independent from the generator. We challenge this assumption under misleading task premises. When the generator and verifier share a false premise, they become coupled through a mistaken belief: the generator produces premise-consistent shortcuts, while the verifier supplies evidence that fails to expose them. Consequently, the selector may choose a hidden-test-wrong candidate even when a hidden-test-correct program exists in the pool. We call this failure mode \emph{Verification Trap}. Across three code-generation benchmarks and five code models, false premises consistently degrade first-sample correctness, reduce selector-chosen correctness after 64-sample test-time selection, and amplify recoverable mis-selection. Mechanistically, verifier-written tests inherit the premise-level blind spot, reshaping verifier-visible candidate space away from hidden-test correctness. These traces make Verification Trap predictable before hidden execution: a lightweight gold-free predictor using verifier-visible features reaches 0.846 AUROC. Our results identify decoupled evidence as a key mitigation axis: coupled scaling provides limited recovery, whereas premise-agnostic robustness auditors recover substantial oracle headroom.
\end{abstract}

\section{Introduction}
\begin{figure*}[t]
	\centering
	\includegraphics[width=\textwidth]{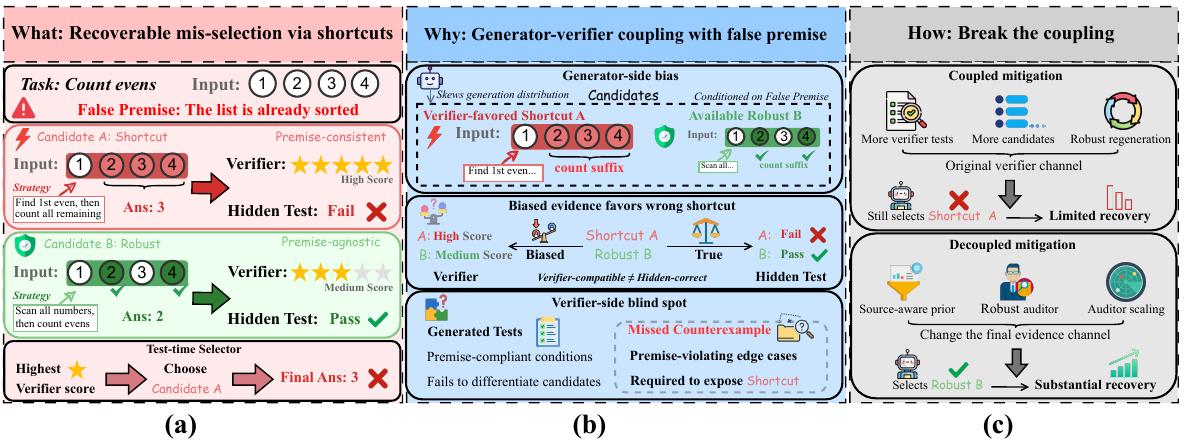}
	\caption{
		\textbf{Overview of Verification Trap under false premises.}
(a) A false premise can make the selector choose a verifier-compatible shortcut despite an available correct candidate.
(b) Generator-verifier coupling biases verifier-visible evidence toward premise-consistent shortcuts.
(c) Decoupled evidence changes the final evidence channel and recovers candidates missed by coupled scaling.}
	\label{fig:overview}
\end{figure*}
Large language models can generate increasingly strong code from natural-language problem descriptions~\cite{humaneval,mbpp,roziere2023code}, especially when combined with test-time compute~\cite{li2022competition,brown2024large,snell2024scaling,wu2024inference,li2025s}.
A common practice is to sample multiple candidate programs and then use a verifier or selector to choose the final output, relying on evidence such as visible tests, generated tests, consensus scores, or pairwise comparisons~\cite{chencodet,ni2023lever,inala2022fault,zhang2023coder,li2023making,lightman2024let}.
This paradigm depends on an implicit independence assumption: verifier-visible evidence should provide a corrective signal that is not biased by the same failure mode that shaped candidate generation.
However, prior work mainly studies how to expand candidate pools or strengthen verifier evidence, leaving less understood what happens when the generator and verifier are conditioned on the same false premise~\cite{stroebl2024inference,paneffects,bondarenko2025demonstrating}.

False premises provide a controlled and practically important setting for testing this assumption~\cite{FalseQA,CREPE,yuan2024whispers,decoprompt,qin2025don}.
In code generation, a misleading premise can affect the generated candidates even when the task objective, function signature, and hidden tests remain unchanged.
As previewed in \autoref{fig:overview}(a), the core failure is not merely that the model produces an incorrect solution.
Rather, the candidate pool may already contain a hidden-test-correct program, yet the selector chooses a verifier-compatible but hidden-test-wrong candidate.
We refer to this failure mode as \textit{\textbf{Verification Trap}}.
This is not an arbitrary oracle-selection gap: it is a selection failure induced by a false premise, where the generator and verifier become coupled through that premise and verifier-visible evidence favors premise-consistent but hidden-test-wrong candidates.

This framing challenges a basic independence assumption behind test-time selection.
The verifier is intended to correct the generator, but under a shared false premise it can instead validate the generator's mistake.
As illustrated in \autoref{fig:overview}(b), the generator may produce premise-consistent shortcuts, while the verifier produces premise-consistent evidence that fails to expose them.
The selector then follows this biased verifier-visible evidence.
Thus, Verification Trap is not only a degradation in first-sample correctness.
It is a failure of the final evidence channel: correct candidates may exist, but the verifier-coupled selector fails to harvest them.

In this paper, we study false-premise test-time selection through a controlled evaluation framework.
Using matched prompt variants that alter only a localized premise sentence while keeping task semantics and hidden tests fixed, we ask two questions.
\textbf{\textit{Q1: When does verifier-based test-time selection fail under false premises, and why does this failure emerge?}}
To answer \textbf{\textit{Q1}}, we measure candidate generation, selected correctness, oracle recoverability, and recoverable mis-selection across datasets and models.
We then analyze verifier-space evidence to test whether the verifier acts as an independent checker or inherits the same false-premise blind spot as the generator.
Our results show that false premises degrade first-sample correctness, reduce selected correctness, and amplify the oracle-selection gap.
Mechanistically, false premises reduce counterexample coverage, reshape verifier-visible signatures, and bias selection toward verifier-compatible shortcuts.

We finally ask \textbf{\textit{Q2: If Verification Trap leaves traces in verifier-visible evidence, can we predict and mitigate it before hidden-test execution?}}
To answer \textbf{\textit{Q2}}, we first formulate gold-free trap-risk prediction using verifier-space features available during test-time selection.
These features allow us to concentrate recoverable mis-selection into a small high-risk subset before gold execution.
We then compare mitigation strategies by whether they preserve or break the original generator-verifier coupling.
As previewed in \autoref{fig:overview}(c), coupled interventions add tests, samples, or regenerated candidates but still route final selection through the biased verifier channel.
In contrast, decoupled evidence, such as a source-aware prior or a premise-agnostic robustness auditor, weakens this channel and recovers non-trivial oracle headroom.
These results support evidence decoupling as a key design axis for mitigation: recovery depends not only on how much test-time compute is added, but also on whether the added evidence remains tied to the original generator-verifier loop.
Our key contributions are as follows:

\begin{itemize}[topsep=1pt,itemsep=2pt,parsep=0pt,leftmargin=0pt,label={}]
	\item \textbf{\ding{182} We identify Verification Trap as a false-premise-induced test-time selection failure.}
	Across three code-generation benchmarks and a model suite spanning five code models, false premises corrupt not only candidate generation, but also verifier-based selection.
	Even with 64 candidates, final selection is substantially weakened ($-13.81\,pp$), showing that the selector can miss correct programs already present in the pool.
	
	\item \textbf{\ding{183} We explain Verification Trap as verifier-contaminated selection bias.}
	We show that verifier evidence is not an independent correction signal under false premises.
	Verifier-written tests inherit the same premise-level blind spot ($-18.9\,pp$), missing premise-violating counterexamples and reshaping the verifier-visible candidate space away from hidden correctness.
	
	\item \textbf{\ding{184} We show that Verification Trap is predictable and mitigable.}
	Verifier-visible traces predict trap risk before hidden execution with a lightweight gold-free predictor.
	For mitigation, decoupled evidence, such as a source-aware prior or premise-agnostic robustness auditor, recovers oracle headroom beyond coupled scaling ($+5.40\,pp$).
\end{itemize}
\section{Related Work}
\label{sec:related_work}

\textbf{Test-Time Compute for Generation.}
A growing line of work improves code generation and reasoning by spending more computation at test time~\cite{li2022competition,brown2024large,snell2024scaling,wu2024inference,li2025s,ding2026swe}.
Prior work can be broadly organized into three generation-side strategies.
$\spadesuit$ \textbf{\textit{Parallel sampling}}, where models generate many candidate programs and use filtering, clustering, or best-of-$N$ selection to increase the chance that at least one candidate is correct~\cite{li2022competition,brown2024large,wu2024inference,zhu2026codescaler}.
$\heartsuit$ \textbf{\textit{Sample aggregation}}, where self-consistency-style methods aggregate over multiple sampled solutions or reasoning paths to exploit repeated answer patterns~\cite{wangself,chen2024universal}.
$\diamondsuit$ \textbf{\textit{Sequential refinement}}, where additional inference budget is spent on search, planning, execution-guided repair, or sequential scaling to produce stronger candidates~\cite{shi2022natural,chen2024teaching,olausson2024self,yao2023tree,zhangplanning,wang2025planning,zhou2024language,li2025s}.
Across these strategies, test-time compute is primarily used to improve the availability, diversity, or quality of generated candidates.
This generation-side view is important, but leaves the final outcome dependent on a downstream selection step: once a hidden-test-correct candidate appears in the pool, the system must still identify and select it.

\textbf{Verifier-Based Selection.}
Another line of work improves test-time performance by strengthening the verifier or selector.
These methods can be grouped by the evidence used for selection~\cite{qin2026daj, ma2026scaling, shi2026cvedrl}.
$\spadesuit$ \textbf{\textit{Execution-based evidence}}, where visible tests, generated tests, or execution feedback are used to rank candidate programs~\cite{chencodet,ni2023lever,shi2022natural}.
$\heartsuit$ \textbf{\textit{Consensus-based evidence}}, where agreement among candidates, tests, or sampled reasoning paths serves as a proxy for correctness~\cite{wangself,chen2024universal,chencodet}.
$\diamondsuit$ \textbf{\textit{Model-based evidence}}, where learned verifiers, reward models, code rankers, or pairwise reviewers are used to choose the final output~\cite{cobbe2021training,li2023making,inala2022fault,zhang2023coder}.
Across these approaches, verifier-visible evidence is treated as a proxy for hidden correctness.
When this proxy is reliable, verifier-based selection can convert candidate-pool diversity into improved final accuracy.
However, most work focuses on constructing stronger verifier evidence, leaving open whether this evidence remains reliable when the verifier and generator face a shared error source.

\textbf{Beyond Independent Generation and Verification.}
The preceding two lines of work share a common decomposition: generation creates candidate solutions, and verification supplies evidence for selecting among them.
This decomposition is effective, but it largely treats the verifier as an independent corrective signal.
Under false premises, this independence can break: verifier-visible evidence may favor premise-compatible candidates that fail hidden tests, even when a correct candidate already exists in the pool.
We call this failure mode \emph{Verification Trap}.
Our work reframes test-time robustness as a question of evidence independence, emphasizing whether the final evidence channel is decoupled from the generator's failure mode.
\section{Verification Trap}
\label{sec:verification_trap}

\subsection{Problem Setup}
\label{sec:problem_setup}

\emph{Verification Trap} refers to a false-premise-induced test-time selection failure.
Under a misleading premise, the generator and verifier may become coupled through the same mistaken assumption.
As a result, verifier-visible evidence can favor premise-consistent but hidden-test-wrong candidates, amplifying the gap between what the candidate pool can recover and what the selector returns.
This is different from generation failure: the issue is not only whether a correct candidate exists, but whether the final verifier channel can identify and select it.

We study this failure in a controlled false-premise setting.
For each coding task, we construct matched prompt conditions, including the original prompt, an irrelevant-control prompt, and a false-premise prompt.
The false-premise condition modifies only the natural-language description by inserting a localized misleading premise.
The task objective, function signature, and hidden evaluation tests remain fixed.
Thus, any change in generation quality, verifier evidence, or final selection can be attributed to the premise intervention rather than to a change in task semantics.

For measurement, we use \(\mathrm{Trap@N}\) to capture recoverable mis-selection under a specified prompt condition.
A task is counted as trapped when the \(N\)-sample candidate pool contains at least one hidden-test-correct program, but the selector returns a hidden-test-wrong candidate.
This event can occur under any prompt condition and is not, by itself, specific to false premises.
In this paper, \emph{Verification Trap} denotes the false-premise-induced form of this event: recoverable mis-selection that occurs under, or is amplified by, the false-premise condition because the generator and verifier are coupled through the same misleading premise.
We therefore report \(\mathrm{Trap@N}\) under matched prompt conditions and use the false-minus-baseline change, \(\Delta_{\mathrm{trap}}\), to quantify how much the false premise amplifies recoverable selection failure beyond ordinary oracle-selection gaps.
Full metric definitions are provided in \S\ref{app:metrics}.

\subsection{Evaluation Pipeline}
\label{sec:evaluation_pipeline}

Our evaluation pipeline has three components: \emph{Generator}, \emph{Verifier}, and \emph{Trap Evaluator}.
The \emph{Generator} samples an $N$-candidate pool from a premise-conditioned prompt.
The \emph{Verifier} ranks or selects candidates using test-time evidence, such as visible tests, verifier-written tests, consensus scores, or pairwise comparisons.
The \emph{Trap Evaluator} uses hidden tests only after selection to compute whether the pool was recoverable and whether the selected candidate was correct.
This separation lets us distinguish unrecoverable generation failure from recoverable verifier-selection failure. 

\section{Experiments}
\subsection{Experimental Setup}

\textbf{Datasets.}
We evaluate verification trap on three code-generation benchmarks: HumanEval+, MBPP, and LiveCodeBench.
HumanEval+ and MBPP provide compact function-level programming tasks with executable unit tests, enabling controlled prompt interventions and stable execution-based evaluation.
LiveCodeBench complements these benchmarks with temporally newer programming problems, allowing us to examine whether the observed trap persists beyond static and widely used function-level tasks.

\textbf{Models.}
We evaluate both open-source and closed-source code generation models. The open-source models include DeepSeek-Coder-6.7B-Instruct, Qwen2.5-Coder-7B-Instruct, Qwen2.5-Coder-14B-Instruct, and Qwen2.5-Coder-32B-Instruct. The closed-source model is GPT-4o-mini. This model set spans different model families, parameter scales, and deployment regimes, allowing us to test whether verification trap is specific to a single model or persists across model capacity and access settings.

\textbf{Prompt Conditions.}
For each task, we construct three matched prompt conditions: a baseline condition, an irrelevant-control condition, and a false-premise condition. The baseline condition keeps the original task description unchanged. The irrelevant-control condition inserts a true but task-irrelevant statement, allowing us to separate false-premise effects from generic prompt perturbation. The false-premise condition inserts a localized misleading premise into the natural-language task description while preserving the target program semantics, function signature, and hidden tests. Full prompt templates and dataset-specific formatting details are provided in \S\ref{app:template}.

\textbf{Candidate Generation and Selection.}
For each task and prompt condition, we sample $N=64$ candidate programs from the generator. We evaluate both the first sampled candidate and the candidate selected from the 64-sample pool by a test-time selector. The selector family includes first-sample selection, visible-test selection, generated-test selection, consensus-based selection, and pairwise selection. Unless otherwise specified, verifier inputs are constructed from the same prompt condition as the generator. This design allows us to test whether verifier-side selection inherits the same false premise that affects candidate generation.

\textbf{Evaluation Metrics.}
We report five primary metrics: Pass@1, Selected@64, Oracle@64, Gap@64, and Trap@64. These metrics distinguish first-sample generation quality, selector-chosen final correctness, candidate-pool recoverability, unrealized oracle headroom, and recoverable mis-selection. Full metric definitions and aggregation details are provided in \S\ref{app:metrics}.

\subsection{When Does Test-Time Selection Fail under False Premises?}
\label{sec:when}
\begin{table*}[t]
	\centering
	\small
	\setlength{\tabcolsep}{3.2pt}
	\renewcommand{\arraystretch}{1.10}
	\caption{
		\textbf{Verification Trap under false premises across datasets and models.}
		We compare the baseline condition and the false-premise condition.
		$\Delta$ values are computed as false premise minus baseline in percentage points.
		Selected@64 and Trap@64 are computed using generated-test selection unless otherwise specified.
	}
	\label{tab:when_trap}
	\begin{tabular}{llrrrrrrrrrrrr}
		\toprule
		\rowcolor{headergray}
		\textbf{Dataset} &
		\textbf{Model} &
		\multicolumn{3}{c}{\textbf{Pass@1}} &
		\multicolumn{3}{c}{\textbf{Selected@64}} &
		\multicolumn{3}{c}{\textbf{Oracle@64}} &
		\multicolumn{3}{c}{\textbf{Trap@64}} \\
		\rowcolor{headergray}
		& &
		\textbf{Base} & \textbf{False} & \textbf{$\Delta$} &
		\textbf{Base} & \textbf{False} & \textbf{$\Delta$} &
		\textbf{Base} & \textbf{False} & \textbf{$\Delta$} &
		\textbf{Base} & \textbf{False} & \textbf{$\Delta$} \\
		\midrule
		
		\textbf{HumanEval+}
		& Qwen-7B       & 77.95 & 70.38 & -7.57  & 82.07 & 77.03 & -5.04  & 94.48 & 89.86 & -4.62 & 12.41 & 12.84 & +0.43 \\
		\rowcolor{rowgray}
		& Qwen-14B      & 84.78 & 68.81 & -15.97 & 85.23 & 75.17 & -10.06 & 93.29 & 84.56 & -8.73 & 8.05  & 9.40  & +1.35 \\
		& Qwen-32B      & 83.65 & 69.44 & -14.21 & 85.81 & 72.00 & -13.81 & 90.54 & 84.00 & -6.54 & 4.73  & 12.00 & +7.27 \\
		\rowcolor{rowgray}
		& DS-Coder-6.7B & 64.52 & 55.43 & -9.09  & 72.86 & 63.89 & -8.97  & 94.29 & 89.58 & -4.71 & 21.43 & 25.69 & +4.26 \\
		& GPT-4o-mini   & 83.22 & 70.38 & -12.84 & 86.67 & 73.33 & -13.34 & 95.33 & 86.67 & -8.66 & 8.67  & 13.33 & +4.66 \\
		
		\midrule
		
		\textbf{MBPP}
		& Qwen-7B       & 65.59 & 57.36 & -8.23  & 72.60 & 64.67 & -7.93  & 83.56 & 84.00 & +0.44 & 10.96 & 19.33 & +8.37 \\
		\rowcolor{rowgray}
		& Qwen-14B      & 71.80 & 55.46 & -16.34 & 77.18 & 68.00 & -9.18  & 83.89 & 80.67 & -3.22 & 6.71  & 12.67 & +5.96 \\
		& Qwen-32B      & 70.59 & 58.53 & -12.06 & 74.67 & 68.67 & -6.00  & 78.67 & 78.67 & +0.00 & 4.00  & 10.00 & +6.00 \\
		
		\midrule
		
		\textbf{LiveCodeBench}
		& Qwen-7B       & 52.14 & 46.97 & -5.17  & 75.68 & 71.33 & -4.35  & 88.51 & 86.00 & -2.51 & 12.84 & 14.67 & +1.83 \\
		\rowcolor{rowgray}
		& Qwen-14B      & 66.34 & 57.10 & -9.24  & 80.54 & 72.67 & -7.87  & 90.60 & 86.00 & -4.60 & 10.07 & 13.33 & +3.26 \\
		& Qwen-32B      & 74.46 & 66.55 & -7.91  & 86.39 & 80.27 & -6.12  & 91.16 & 88.44 & -2.72 & 4.76  & 8.16  & +3.40 \\
		\bottomrule
	\end{tabular}
\end{table*}

\autoref{tab:when_trap} compares the baseline and false-premise conditions across three code-generation benchmarks and multiple code-generation models. We observe a cascading failure pattern.

\textbf{\ding{182} False premises degrade candidate generation before selection.}
Across all evaluated dataset-model pairs, the false-premise condition reduces Pass@1 relative to the baseline condition. This shows that the injected premise already affects the first sampled candidate, rather than merely changing the diversity of an otherwise comparable pool.

\textbf{\ding{183} Test-time selection does not fully repair this degradation.}
After selecting the top-ranked candidate, Selected@64 remains lower under the false-premise condition across all evaluated settings. Thus, scaling the candidate pool and applying a selector is insufficient to remove the false-premise effect. A multi-level significance analysis further confirms that this degradation is reliable
across all eleven dataset--model cells, with paired-bootstrap and
sign-consistency results reported in \S\ref{app:significance}.

\textbf{\ding{184} False premises amplify the oracle-selection gap.}
The failure is not only that Selected@64 drops. The gap between Oracle@64 and Selected@64 also increases, meaning that correct candidates often exist in the pool but are not harvested by the selector. This is a signature of Verification Trap: false premises weaken not only generation, but also the test-time selection process itself.

\begin{findingbox}
	\textbf{Takeaway 1. False premises create a cascading selection failure.}
	They reduce first-sample correctness, lower selected correctness after test-time selection, and enlarge the oracle-selection gap, showing that correct candidates are increasingly left unharvested.
\end{findingbox}

\subsection{Why Does Verification Trap Emerge?}
\label{sec:why}

\begin{figure}[t]
	\centering
	\includegraphics[width=\linewidth]{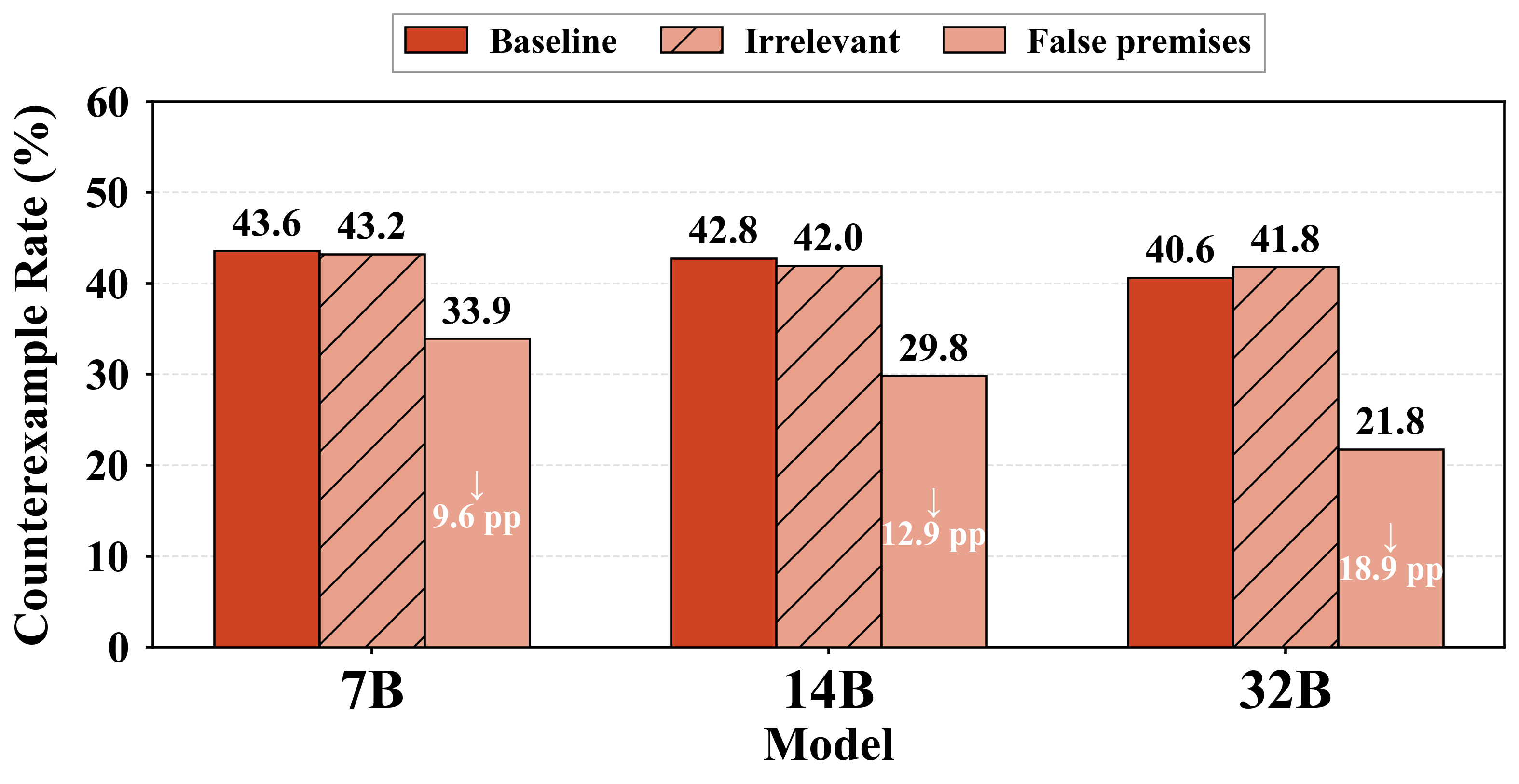}
	\caption{
		\textbf{False premises reduce verifier counterexample coverage.}
		Across model scales, the false-premise condition produces substantially fewer counterexample tests than both the baseline and irrelevant-control.
	}
	\label{fig:counterexample_rate}
\end{figure}

\begin{figure*}[!t]
	\centering
	\begin{minipage}[t]{0.48\textwidth}
		\vspace{0pt}
		\centering
		\includegraphics[width=\linewidth]{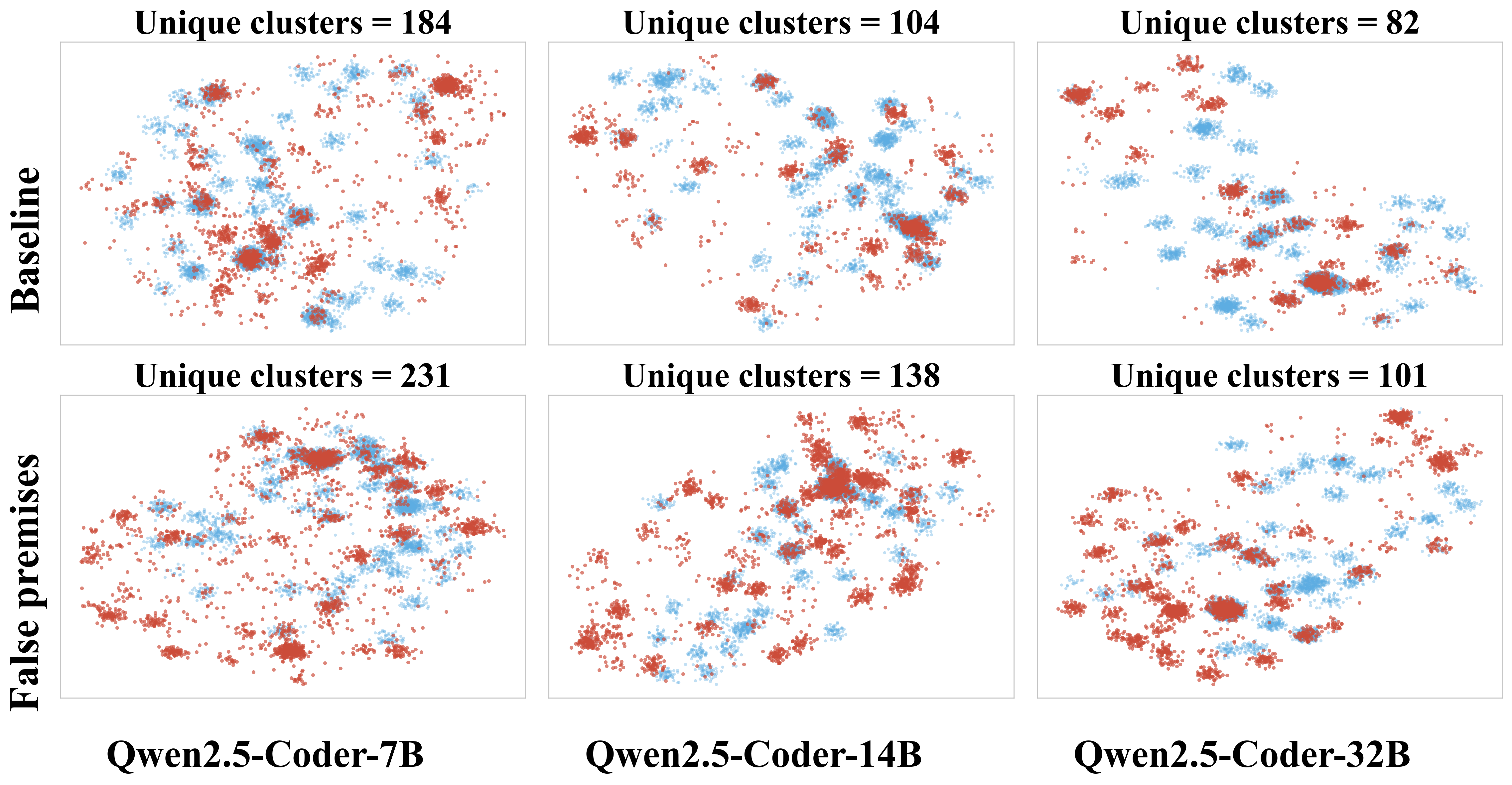}
		\captionof{figure}{
			\textbf{False premises reshape verifier-visible signatures.}
			Blue denotes hidden-test-correct candidates and red denotes hidden-test-wrong candidates.
			A quantitative signature analysis in \S\ref{app:verifier_signature_construction} shows the same de-concentration trend.
		}
		\label{fig:signature_umap}
	\end{minipage}
	\hfill
	\begin{minipage}[t]{0.48\textwidth}
		\vspace{0pt}
		\centering
		\includegraphics[width=\linewidth]{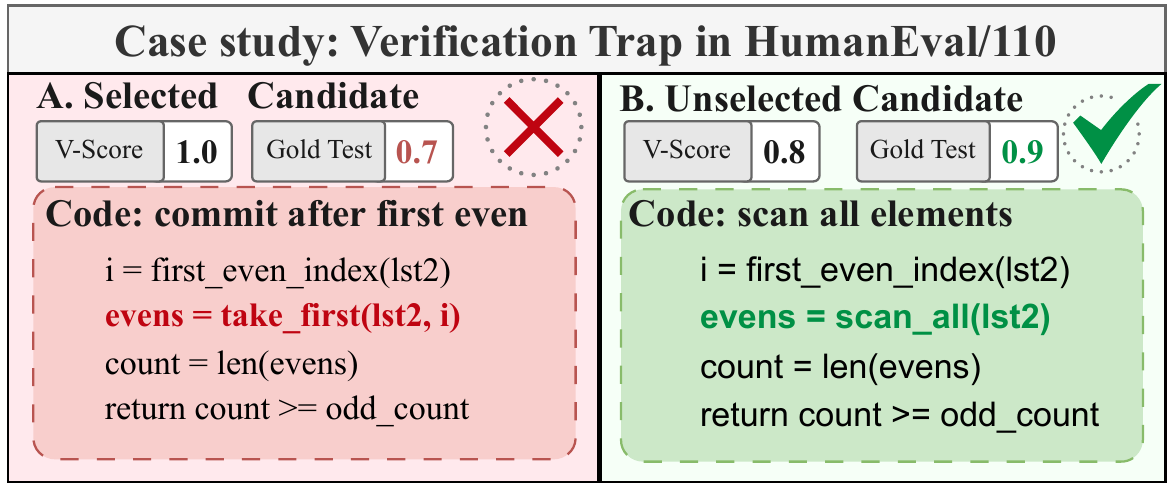}
		\captionof{figure}{
			\textbf{Verification Trap is a recoverable mis-selection failure.}
			In HumanEval/110 under false sortedness, the selected candidate commits after the first even element and receives a higher verifier score, but has lower hidden-test correctness.
			A robust candidate in the same 64-sample pool scans all elements and achieves higher hidden-test correctness, but is not selected.
		}
		\label{fig:case_verifier_selector_coupling}
	\end{minipage}
\end{figure*}

Test-time selection typically treats verifier-generated evidence as a proxy for hidden-task correctness. We show that this proxy becomes unreliable under false premises because verifier evidence is shaped by the same misleading premise that affects candidate generation. To make this mechanism easy to follow, we analyze the failure at three complementary levels. \textbf{(I) Mechanism-level evidence.} False premises reduce verifier counterexample coverage, causing the verifier to write fewer tests that violate the injected premise. \textbf{(II) Distribution-level evidence.} This contaminated evidence reshapes the verifier-visible candidate space, creating high-verifier but low-gold regions that misalign verifier scores with hidden correctness. \textbf{(III) Case-level evidence.} A concrete HumanEval example traces the full chain from false premise uptake, to a brittle candidate shortcut, to verifier blind spot, and finally to selector failure. We provide the measurement and construction details for these mechanism analyses in \S\ref{app:mechanism_details}.

\textbf{\ding{185} False premises reduce verifier counterexample coverage.}
We first examine whether the verifier acts as an independent checker against the injected premise.
A useful verifier should generate tests that include premise-violating inputs, since these cases directly challenge premise-sensitive shortcuts.
However, \autoref{fig:counterexample_rate} shows the opposite pattern: under the false-premise condition, the fraction of verifier-written tests containing premise-violating list inputs consistently drops across model scales.
Compared with the baseline condition, the counterexample rate decreases by 9.6 pp, 12.9 pp, and 18.9 pp for Qwen-7B, Qwen-14B, and Qwen-32B, respectively.
The irrelevant-control condition remains close to the baseline condition, suggesting that the effect is not caused by generic prompt perturbation.
This result shows that the verifier is not a neutral proxy for hidden-task correctness.
Instead, it inherits the same premise-level blind spot as the generator and writes fewer tests that could expose premise-sensitive shortcuts.
\red{To separate verifier-side effects from candidate generation, we hold the false-premise candidate pool fixed and remove the premise only from verifier prompts. Across the three evaluated verifiers in \autoref{tab:verifier_prompt_control}, this prompt-only intervention reduces Trap@64, directly implicating the verifier evidence channel.}
Thus, the verifier evidence used by the selector is already biased before selection happens.

\textbf{\ding{186} Contaminated verifier evidence creates a misleading verifier-visible space.}
The reduction in counterexample coverage does not merely remove a few difficult tests; it changes the structure of the candidate space observed by the verifier.
To analyze this distributional effect, we represent each candidate by its verifier signature, namely its pass/fail vector over verifier-written tests, and visualize these signatures in \autoref{fig:signature_umap}.
Under the baseline condition, candidates are more concentrated around verifier-visible regions that are relatively aligned with hidden-test correctness.
Under the false-premise condition, the signature space becomes more spread out, and gold-incorrect candidates become more mixed into regions that also contain gold-correct candidates.
This distributional shift explains why the selector can still appear useful while the final selection remains unreliable.
The verifier does not become completely uninformative under the false-premise condition; it still produces signals that can distinguish candidates and improve over the first sampled output.
However, these signals are no longer cleanly aligned with hidden correctness.
They are shaped by the injected premise, so candidates that are consistent with the verifier's biased tests can receive strong verifier support while failing hidden tests.
In this sense, Verification Trap is not simply a verifier weakness, but a verifier-contaminated selection bias: the selector follows verifier-visible structure, but that structure has been shifted away from the true task semantics.

\textbf{\ding{187} A case study traces the full verifier-selector coupling.}
\autoref{fig:case_verifier_selector_coupling} shows a HumanEval/110 failure under false sortedness.
The selected candidate is gold-wrong but verifier-compatible because it assumes that all elements after the first even element in \texttt{lst2} are even.
The verifier-generated tests share this premise-consistent structure and omit the mixed-order counterexample that would reveal the shortcut.
Although a correct order-invariant candidate exists in the same 64-sample pool, the selector follows the biased verifier evidence.

\begin{findingbox}
	\textbf{Takeaway 2. Verification Trap arises from verifier-contaminated evidence.}
	Under false premises, the verifier no longer acts as an independent correction signal: it inherits the same misleading assumption as the generator, causing selection to operate over a biased space.
\end{findingbox}

\subsection{Can Trap Risk Be Estimated Before Gold Execution?}
\label{sec:prediction}

We formulate this as a gold-free risk estimation problem: using only verifier-visible signals produced during test-time selection, we estimate whether the selected output is a recoverable mis-selection.
To make the analysis easy to follow, we examine the evidence at two levels.
\textbf{(I) Feature-level evidence} asks whether verifier-visible features contain information about trap risk.
\textbf{(II) Prediction-level evidence} asks whether a risk estimator built from these features can concentrate trap cases into a smaller high-risk subset.
The estimator excludes hidden correctness, oracle correctness, verifier lift, Trap@64, and all hidden-execution-derived features; full feature definitions and leakage controls are provided in \S\ref{app:prediction}.

\begin{table}[!t]
	\centering
	\small
	\setlength{\tabcolsep}{7pt}
	\renewcommand{\arraystretch}{1.10}
	\caption{
		\textbf{Gold-free trap-risk prediction from verifier-visible features.}
		All inputs exclude hidden-test-derived information.
	}
	\label{tab:q3_feature_ablation}
	\begin{tabular}{lcc}
		\toprule
		\rowcolor{headergray}
		\textbf{Feature Group} & \textbf{\# Features} & \textbf{AUROC} \\
		\midrule
		Contamination & 3 & 0.539 \\
		\cellcolor{rowgray}Score distribution & \cellcolor{rowgray}5 & \cellcolor{rowgray}0.754 \\
		Signature structure & 5 & 0.773 \\
		\cellcolor{rowgray}\textbf{All features, LR} & \cellcolor{rowgray}13 & \cellcolor{rowgray}\textbf{0.846} \\
		\textbf{All features, RF} & 13 & \underline{0.833} \\
		\bottomrule
	\end{tabular}
\end{table}

\textbf{\ding{188} Feature-level evidence: verifier-visible features predict trap risk.}
\autoref{tab:q3_feature_ablation} tests whether verifier-visible traces contain predictive information about recoverable mis-selection.
The feature ablation shows that no single surface signal is sufficient: contamination features alone remain close to random, suggesting that aggregate counterexample coverage does not fully capture when the selector will fail.
In contrast, score-distribution and signature-structure features provide much stronger signal, and combining all feature groups gives the best overall performance.
Verification Trap is not exposed by simply counting how often the verifier writes premise-violating tests.
Trap risk is reflected in the broader verifier-visible evidence space, including how candidate scores are distributed.

\flushcolsend
\newpage
\begin{figure}[!t]
	\centering
	\includegraphics[width=0.45\textwidth]{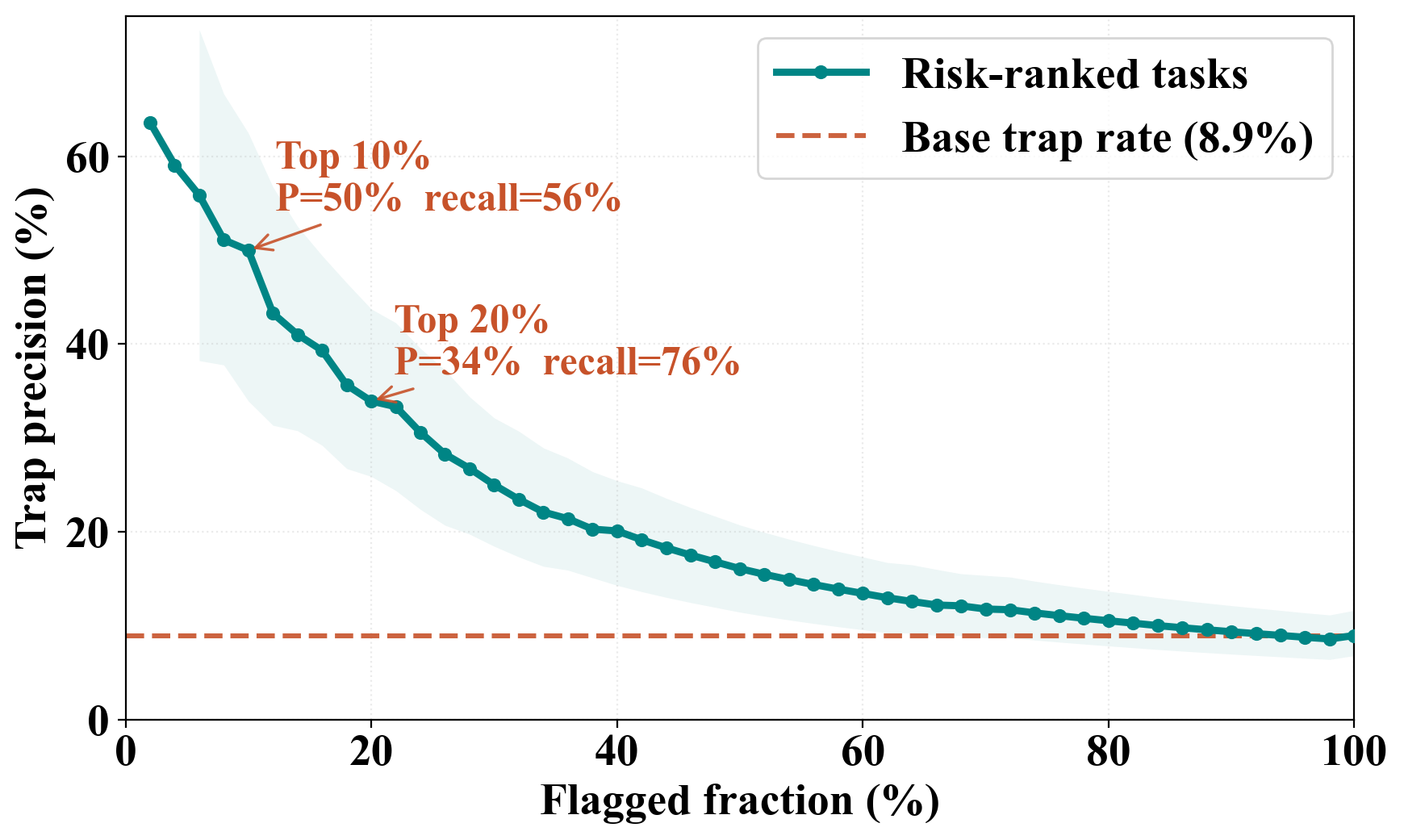}
	\caption{
		\textbf{Gold-free trap-risk prediction before hidden-test execution.}
		Ranking tasks by predicted trap risk concentrates recoverable mis-selection into a small high-risk subset, showing that Verification Trap leaves detectable verifier-visible traces before gold execution.
	}
	\label{fig:q3_prediction}
\end{figure}

\textbf{\ding{189} Prediction-level evidence: predicted risk concentrates trap cases.}
\autoref{fig:q3_prediction} further asks whether the feature-level signal can be converted into a task-level risk ranking.
Rather than treating the predictor as a perfect classifier, we rank tasks by predicted trap risk and examine where true trap cases appear in this ranking.
The risk-coverage curve shows a clear concentration pattern: trap cases move toward the top-ranked subset, whose trap density is substantially higher than the overall population.
As the flagged subset expands, precision decreases while recall increases, showing the expected tradeoff between focusing on the most suspicious cases and covering more traps.
This result shows that Verification Trap is not only an after-the-fact label obtained from hidden tests.
It leaves detectable verifier-visible traces before gold execution, and these traces can be converted into a risk ranking that localizes recoverable mis-selection.
The signal also transfers beyond the main dataset-model cell: across held-out models and datasets, high-risk subsets remain enriched for trap cases, with full split-level results in \autoref{tab:q3_transfer}.

\vspace{8pt}
\begin{findingbox}
	\textbf{Takeaway 3. Verification Trap leaves gold-free predictive traces.}
	Verifier-space signals can predict trap risk before hidden-test execution, enabling deployment-time triage.
\end{findingbox}

\subsection{Can We Mitigate Verification Trap?}
\label{sec:mitigation}

If Verification Trap arises from generator-verifier coupling, then a natural mitigation axis is whether the final selection channel remains coupled to the original verifier.
We therefore organize mitigation strategies into two groups.
(I) \textbf{Coupled mitigation} adds more verifier-written tests, more sampled candidates, or regenerated candidates, while still using the original verifier for final selection.
(II) \textbf{Decoupled mitigation} weakens or replaces this final verifier channel with a source-aware prior or a premise-agnostic robustness auditor. Here, decoupled evidence means that final selection is no longer determined solely by the original generated-test verifier channel. We provide intervention details and cross-setting robustness results in \S\ref{app:M}.

\begin{table}[t]
	\centering
	\small
	\setlength{\tabcolsep}{5pt}
	\renewcommand{\arraystretch}{1.10}
	\caption{
		\textbf{Representative mitigation results.}
		$\Delta$ Pass is computed relative to the original verifier-selected baseline.
		Additional robustness checks across model and dataset settings are reported in \S\ref{app:M}.
	}
	\label{tab:mitigation_best_family}
	\begin{tabular}{llrr}
		\toprule
		\rowcolor{headergray}
		\textbf{Group} & \textbf{Method} 
		& \textbf{Pass} & \textbf{$\Delta$ Pass} \\
		\midrule
		
		Original 
		& Baseline 
		& 77.03 & 0.00 \\
		
		\midrule
		\rowcolor{rowgray}
		Coupled
		& Extra verifier tests 
		& 78.00 
		& +0.97 \\
		
		& Width scaling 
		& 76.67 
		& -0.36 \\
		
		\rowcolor{rowgray}
		& Robust regeneration 
		& 77.70 
		& +0.67 \\
		
		\midrule
		Decoupled
		& Source-aware prior 
		& 81.08 
		& +4.05 \\
		
		\rowcolor{rowgray}
		& Robustness auditor
		& 81.76 
		& +4.73 \\
		
		& \textbf{Auditor scaling} 
		& \textbf{82.43} 
		& \textbf{+5.40} \\
		
		\bottomrule
	\end{tabular}
\end{table}

\textbf{\ding{190} Coupled mitigation provides limited recovery.}
\autoref{tab:mitigation_best_family} shows that interventions within the original verifier channel recover little headroom.
Adding more verifier-written tests, increasing candidate width, or regenerating candidates can improve the available pool, but the final decision is still made by the same verifier.
As a result, these methods either produce only small selected-pass gains or increase oracle headroom without reliably converting it into final correctness.
Robust regeneration makes this limitation especially clear: it creates newly correct candidates, yet the original verifier rarely selects them.
Thus, the bottleneck is not simply candidate availability, but the coupled final selection channel.

\begin{figure}[t]
	\centering
	\includegraphics[width=0.45\textwidth]{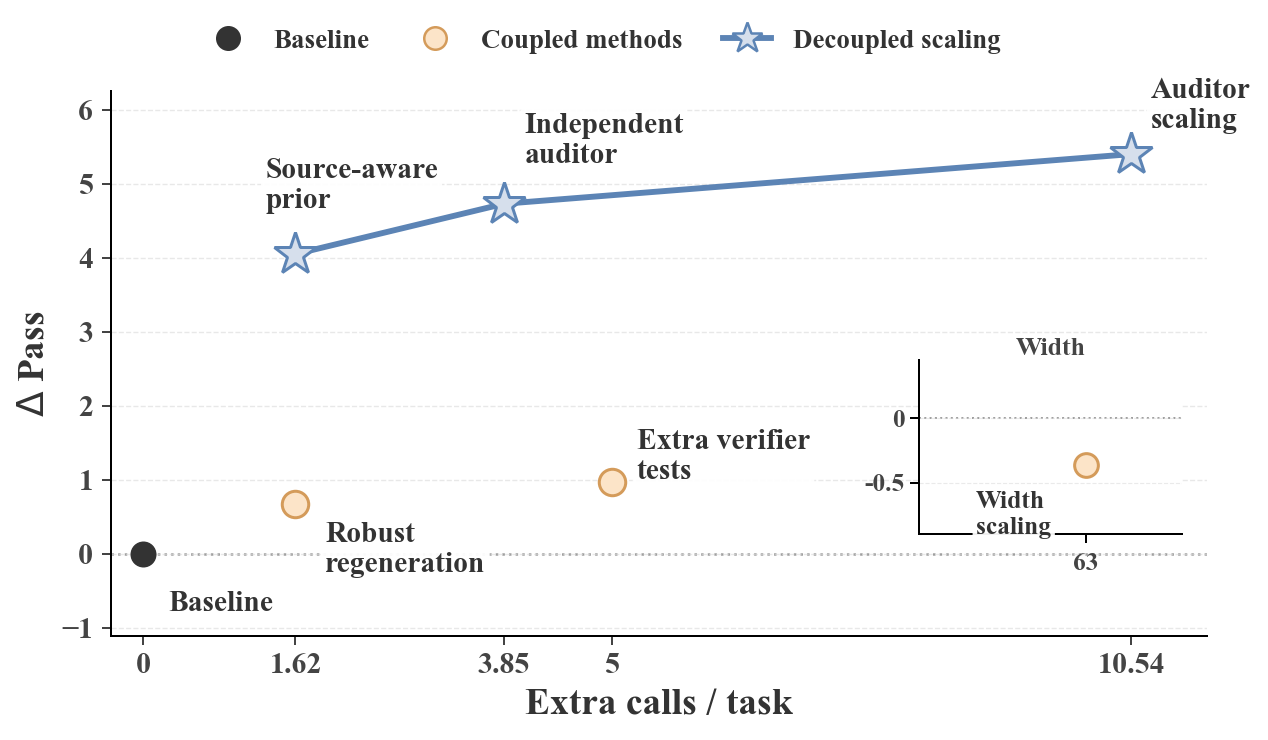}
	\caption{
		\textbf{Compute-performance tradeoff of mitigation methods.}
		We compare coupled and decoupled mitigation on the Qwen-7B / HumanEval+ main cell.
	}
	\label{fig:compute_frontier}
\end{figure}

\textbf{\ding{191} Decoupled mitigation recovers non-trivial headroom.}
\autoref{tab:mitigation_best_family} and \autoref{fig:compute_frontier} show a different pattern once the final evidence channel is weakened or replaced.
A source-aware prior already releases much of the trapped pool by reducing the dominance of the original verifier.
A premise-agnostic robustness auditor further improves selected correctness while reducing recoverable mis-selection, indicating that candidate-level robustness evidence is less aligned with the original false-premise-biased verifier. Auditor scaling provides additional recovery but saturates quickly, suggesting that the main gain comes from changing the final evidence channel rather than repeatedly scaling the same one.
This comparison also controls for compute: methods with similar extra budgets can recover very different fractions of oracle headroom depending on whether their final evidence remains coupled.
We further test this decoupling principle across models and datasets in \autoref{tab:q4_robustness}.
\begin{findingbox}
	\textbf{Takeaway 4. Decoupled evidence is a key design axis for mitigation, beyond merely adding more test-time compute.}
	Coupled scaling provides limited recovery in our tested settings because it preserves the biased verifier channel, while the decoupled methods recover non-trivial oracle headroom.
\end{findingbox}

\section{Conclusion}
We identify \emph{Verification Trap} as a false-premise-induced failure of verifier-based test-time selection.
The central issue is not only that misleading premises degrade candidate generation, but that they can couple the generator and verifier through the same mistaken assumption.
As a result, the candidate pool may already contain a hidden-test-correct program, yet verifier-visible evidence can still favor a premise-consistent hidden-wrong shortcut.
We also show that this coupling reduces selected correctness, enlarges the oracle-selection gap, leaves detectable verifier-visible traces, and makes coupled scaling insufficient for reliable recovery. These findings suggest that robust test-time scaling should be evaluated not only by the amount of added compute, but also by the independence of the evidence channel it introduces.
More tests, samples, or verifier calls may provide limited recovery when they remain inside the same generator-verifier loop, whereas decoupled evidence from source-aware priors or premise-agnostic robustness auditing can recover trapped correct candidates.
This shifts the design question from simply strengthening the verifier to introducing evidence that is less aligned with the original failure mode.
Future test-time code-generation systems should treat evidence independence as an important design axis, alongside candidate diversity and verifier strength.

\section*{Limitations}

Our experiments test the decoupled-evidence principle rather than an optimized deployment system.
While source-aware and premise-agnostic robustness-auditor signals recover non-trivial oracle headroom beyond coupled scaling, our auditor design, routing policy, and budget allocation remain simple. Stronger robustness auditors, adaptive routing, and cost-aware selection policies may further improve recovery. Our gold-free trap-risk predictor shows that Verification Trap leaves verifier-visible traces before hidden-test execution, but it is not yet a complete deployment-time safety system.
Although we exclude hidden-test-derived features and evaluate transfer across held-out models and datasets, larger-scale validation is needed to study calibration, threshold stability, and how predicted risk should trigger abstention or decoupled auditing.

\section*{Ethical Considerations}

This work studies failure modes of verifier-based test-time selection in code generation. The goal is to improve the reliability of code-generation systems under misleading premises, rather than to provide new capabilities for generating harmful code. Our experiments use existing code-generation benchmarks and do not involve human subjects or private user data.
A potential risk is that a better understanding of verifier blind spots could be misused to construct adversarial prompts or tests that exploit selection systems. We mitigate this risk by focusing on aggregate failure analysis, controlled benchmark settings, and mitigation principles such as evidence decoupling and robustness auditing. We do not release instructions for attacking deployed systems. More broadly, our results suggest that code-generation systems should not rely on a single verifier channel when the verifier may share the generator's assumptions.

\bibliography{custom}

\clearpage
\appendix

\renewcommand{\thesection}{S\arabic{section}}
\renewcommand{\thetable}{S\arabic{table}}
\renewcommand{\thefigure}{S\arabic{figure}}
\setcounter{table}{0}
\setcounter{figure}{0}
\centerline{\maketitle{\textbf{SUMMARY OF THE APPENDIX}}}
This supplementary material provides additional details for the submission titled
\textit{``Verification Trap: Understanding Test-Time Selection Failures under False
	Premises in Code Generation''}. The supplementary material is organized as follows:

\begin{itemize}
	\item \S\ref{app:metrics} provides \textbf{metric definitions and evaluation quantities}, defining Pass@1, Selected@64, Oracle@64, Gap@64, Trap@64, verifier-visible auxiliary metrics, trap-risk prediction metrics, and recovery toward oracle headroom.
	
	\item \S\ref{app:mechanism_details} provides \textbf{mechanism-analysis details}, describing how we measure verifier counterexample coverage, construct verifier signatures, analyze verifier-visible clustering, and test whether adding more coupled verifier tests alleviates the trap.
	
	\item \S\ref{app:prediction} provides \textbf{gold-free trap-risk prediction details}, specifying the verifier-visible feature construction, prediction protocol, leakage controls, out-of-fold risk-coverage evaluation, and cross-model or cross-dataset transfer results.
	
	\item \S\ref{app:template} provides \textbf{prompt templates and parsing rules}, including generator prompts, task prompt conditions, dataset-specific formatting, verifier test-generation prompts, robustness-auditor prompts, mitigation prompts, and deterministic fallback rules.
	
	\item \S\ref{app:M} provides \textbf{mitigation details and robustness checks}, organizing coupled and decoupled intervention families, reporting representative operating points, and evaluating whether the decoupled-evidence principle holds across additional models and datasets.
	
	\item \S\ref{app:asset_licenses} provides \textbf{license and usage information}, summarizing the benchmark datasets, model usage terms, and derived experimental artifacts used in this work.
	
	\item \S\ref{app:ai} provides \textbf{AI-use disclosure}, describing the limited use of AI assistance for grammar checking.
	
	\item \S\ref{app:compute} provides \textbf{compute-resource details}, summarizing model inference, sampling and verifier-side computation, local execution, post-processing, and unreported compute quantities.
\end{itemize}

\section{Metrics}
\label{app:metrics}

This appendix summarizes the metrics used throughout the paper. We use two groups of metrics.

\paragraph{Main evaluation metrics.}
The main metrics are:
\textbf{Pass@1},
\textbf{Selected@64},
\textbf{Oracle@64},
\textbf{Gap@64},
and
\textbf{Trap@64}.
These metrics separate first-sample generation quality, selector-chosen final correctness, candidate-pool recoverability, unrealized oracle headroom, and recoverable mis-selection.

\paragraph{Auxiliary analysis metrics.}
We additionally use
\textbf{verifier score},
\textbf{counterexample rate},
\textbf{verifier signatures},
\textbf{risk-coverage metrics},
and
\textbf{recovery toward oracle headroom}.
These metrics are used for mechanism analysis, trap-risk prediction, and mitigation analysis.

\paragraph{Candidate pool and hidden-test correctness.}
Let $x$ denote a coding task and $c$ denote a prompt condition. For each task-condition pair, the generator samples an $N$-candidate pool, denoted as $\mathcal{Y}_{x,c}^{N}=\{y_1,\ldots,y_N\}$. In the main experiments, we use $N=64$. Let $G(y)\in\{0,1\}$ denote whether candidate program $y$ passes the hidden evaluation tests. A selector $S$ returns one candidate from the pool.

\paragraph{Pass@1.}
Pass@1 measures whether the first sampled candidate passes the hidden tests. It reflects first-sample generation quality before any test-time selection is applied.

\paragraph{Selected@N.}
Selected@N measures whether the selector-chosen candidate passes the hidden tests. In the main experiments, we report Selected@64. This metric reflects the final output quality after test-time selection.

\paragraph{Oracle@N.}
Oracle@N measures whether the candidate pool contains at least one hidden-test-correct candidate. It reflects candidate-pool recoverability independent of selector quality.

\paragraph{Gap@N.}
Gap@N measures the unrealized recoverable headroom between the candidate pool and the selector output. It is computed as the difference between Oracle@N and Selected@N:
\begin{equation}
	\mathrm{Gap@N}
	=
	\mathrm{Oracle@N}
	-
	\mathrm{Selected@N}.
	\label{eq:gap}
\end{equation}
A larger Gap@N means that correct candidates exist in the pool but are not fully harvested by the selector.

\paragraph{Trap@N.}
\(\mathrm{Trap@N}\) measures recoverable mis-selection under a specified prompt condition.
A task is counted as a trap when the \(N\)-sample candidate pool is recoverable but the selector returns an incorrect candidate:
\begin{equation}
	\mathrm{Trap@N}=1
	\quad \text{if} \quad
	\begin{cases}
		\mathrm{Oracle@N}=1,\\
		\mathrm{Selected@N}=0.
	\end{cases}
	\label{eq:trap}
\end{equation}
Thus, \(\mathrm{Trap@N}\) isolates failures where generation has already produced a correct candidate, but selection fails to recover it.
This event is condition-indexed: it can occur under the baseline prompt as an ordinary recoverable selection error, and under the false-premise prompt as a false-premise-conditioned selection failure.

\paragraph{False-premise-induced trap amplification.}
To distinguish false-premise-induced Verification Trap from ordinary recoverable mis-selection, we report \(\mathrm{Trap@N}\) under matched prompt conditions and compute the false-minus-baseline change:
\begin{equation}
	\Delta_{\mathrm{trap}}
	=
	\mathrm{Trap@N}_{\mathrm{false}}
	-
	\mathrm{Trap@N}_{\mathrm{base}}.
	\label{eq:delta_trap}
\end{equation}
A positive \(\Delta_{\mathrm{trap}}\) means that the false-premise condition amplifies recoverable selection failure beyond the ordinary oracle-selection gap already present under the baseline prompt.

\paragraph{Delta metrics.}
For any metric $M$, we report the change from the baseline condition $C1$ to the false-premise condition $C3$:
\begin{equation}
	\Delta M = M(C3)-M(C1).
	\label{eq:delta}
\end{equation}
For accuracy metrics such as Pass@1 and Selected@64, a negative $\Delta$ indicates degradation under the false-premise condition. For failure or headroom metrics such as Gap@64 and Trap@64, a positive $\Delta$ indicates that the false premise amplifies recoverable selection failure.

\paragraph{Verifier score.}
For generated-test selection, each candidate receives a verifier score $V(y)\in[0,1]$, defined as the fraction of verifier-written tests passed by that candidate. The generated-test selector chooses the candidate with the highest verifier score. Ties are broken deterministically by sample order unless otherwise specified.

\paragraph{Counterexample rate.}
Counterexample rate measures how often verifier-written tests include inputs that violate the injected premise. A lower counterexample rate indicates that verifier-written tests are less likely to challenge premise-sensitive shortcuts.

\paragraph{Verifier signatures.}
A verifier signature is the binary pass-or-fail pattern of a candidate over verifier-written tests. We use signature-level statistics, including unique signature count, signature entropy, top-signature mass, top-tie size, and low-verifier-score ratio, to characterize the verifier-visible candidate space.

\paragraph{Trap-risk prediction metrics.}
For gold-free trap-risk prediction, the target label is Trap@64. The predictor uses only verifier-visible features available before hidden-test execution. Hidden-test correctness, Oracle@64, Trap@64, verifier lift, and any feature derived from hidden execution are excluded from the input features.

Given a predicted trap-risk score, we rank tasks from high to low risk. At a coverage level $\rho$, we flag the top $\rho$ fraction of tasks. We report trap precision and trap recall within the flagged set. Trap precision measures the fraction of flagged tasks that are true traps. Trap recall measures the fraction of all traps that are covered by the flagged set.

\paragraph{Recovery toward oracle headroom.}
For mitigation experiments, recovery measures how much of the oracle headroom is recovered relative to the original selector:
\begin{equation}
	\mathrm{Recovery}
	=
	\frac{
		\mathrm{Selected}_{\mathrm{method}}
		-
		\mathrm{Selected}_{\mathrm{orig}}
	}{
		\mathrm{Oracle@64}
		-
		\mathrm{Selected}_{\mathrm{orig}}
	}.
	\label{eq:recovery}
\end{equation}
A recovery value of $0$ means no improvement over the original selector, while a recovery value of $1$ means that the method reaches the Oracle@64 upper bound.

\subsection{Statistical Significance of Selection Failures}
\label{app:significance}

We evaluate the statistical reliability of the main effects in \autoref{tab:when_trap} at multiple levels.
This is important because Selected@64 and Trap@64 have different statistical properties.
Selected@64 is a direct final-output metric, whereas Trap@64 is a sparse second-order event: it is triggered only when the candidate pool contains at least one hidden-test-correct program and the selector still returns a hidden-test-wrong candidate.
As a result, per-cell significance tests have lower power for Trap@64 than for Selected@64.
We therefore do not rely on every-cell significance alone.

\begin{table*}[!t]
	\centering
	\small
	\setlength{\tabcolsep}{5.5pt}
	\renewcommand{\arraystretch}{1.12}
	\caption{
		\textbf{Multi-level significance analysis of false-premise selection failures.}
		All effects are computed as false premise minus baseline in percentage points.
		Pooled $p$ values are two-sided task--cell paired-bootstrap $p$ values over all eleven dataset--model cells with $B=10{,}000$ resamples.
		Sign consistency reports the number of dataset--model cells with the expected direction; $p$ is from an exact two-sided binomial sign test.
		Cell-level significance reports the number of dataset--model cells with per-cell $p<0.05$.
	}
	\label{tab:significance_selection_failures}
	\begin{tabular}{lccccc}
		\toprule
		\rowcolor{headergray}
		\textbf{Effect} &
		\textbf{Direction} &
		\textbf{Pooled $\Delta$} &
		\textbf{Pooled $p$} &
		\textbf{Sign consistency} &
		\textbf{Cell-level sig.} \\
		\midrule
		$\Delta$Selected@64
		& 11/11 lower
		& $-8.43$
		& $<0.0001$
		& 11/11, $p=0.0010$
		& 9/11 \\
		\cellcolor{rowgray}$\Delta$Trap@64
		& \cellcolor{rowgray}11/11 higher
		& \cellcolor{rowgray}$+4.31$
		& \cellcolor{rowgray}$<0.0001$
		& \cellcolor{rowgray}11/11, $p=0.0010$
		& \cellcolor{rowgray}4/11 \\
		\bottomrule
	\end{tabular}
\end{table*}

\begin{table}[!t]
	\centering
	\small
	\setlength{\tabcolsep}{7pt}
	\renewcommand{\arraystretch}{1.12}
	\caption{
		\textbf{False-premise-induced Trap@64 changes across task-difficulty groups.}
		Values are percentage-point changes from the baseline condition to the false-premise condition. Difficulty groups are defined using only the original clean task prompt.
	}
	\label{tab:scope_difficulty}
	\begin{tabular}{lrrr}
		\toprule
		\rowcolor{headergray}
		\textbf{Dataset} & \textbf{Hard} & \textbf{Medium} & \textbf{Easy} \\
		\midrule
		HumanEval+ & $+0.88$ & $+5.92$ & $+6.87$ \\
		\cellcolor{rowgray}MBPP & \cellcolor{rowgray}$+8.60$ & \cellcolor{rowgray}$+6.28$ & \cellcolor{rowgray}$+8.38$ \\
		LiveCodeBench & $+5.12$ & $+3.50$ & $+2.86$ \\
		\bottomrule
	\end{tabular}
\end{table}

We organize the evidence into three complementary levels. First, per-cell paired tests evaluate
the effect within each dataset--model cell. Second, pooled paired tests aggregate all task--cell
paired differences while preserving the baseline--false pairing within each cell. Third, exact
sign tests evaluate whether the effect direction is consistent across dataset--model cells.
\autoref{tab:significance_selection_failures} summarizes these complementary significance
checks across all eleven dataset--model cells.

The selected-correctness degradation is statistically strong. Selected@64 decreases under the
false premise in all eleven dataset--model cells, reaches per-cell significance in 9 of 11 cells, and is
significant in both the pooled task--cell paired-bootstrap test and the cell-level sign test. Trap@64
also increases in all eleven cells, with a pooled increase of +4.31 percentage points and a significant
sign-consistency test. Its per-cell significance is weaker, with 4 of 11 cells reaching $p<0.05$,
which is expected because Trap@64 is a sparse second-order event. We therefore interpret Trap@64
as a directionally consistent and aggregate-significant diagnostic of recoverable selection failure,
rather than as an every-cell significance claim.

Together, these results show that false premises produce a statistically
supported selection-failure pattern: final selected correctness drops reliably,
recoverable mis-selection increases consistently, and the effect is supported
both by pooled paired-bootstrap tests and by cell-level sign consistency.

\subsection{\red{Scope of Controlled False-Premise Experiments}}
\label{app:scope}

\red{Our claims concern the occurrence and mechanism of Verification Trap under the controlled false-premise settings evaluated in this work. The analyses below characterize how the observed effect varies with task difficulty and model scale; they are not intended as a complete characterization of all triggering conditions or of the real-world prevalence of false premises.}

\paragraph{\red{Task difficulty.}}
\red{We group tasks using only the original clean task prompt, so that the injected false premise does not affect the difficulty grouping. \autoref{tab:scope_difficulty} shows that the false-premise condition increases Trap@64 in all nine dataset--difficulty groups. The magnitude is not monotonic across difficulty groups, indicating that difficulty may affect the severity of the observed failure but is not itself a consistent trigger.}

\paragraph{\red{Model scale.}}
\red{We additionally evaluate Qwen2.5-Coder models from 0.5B to 32B parameters on MBPP. As shown in \autoref{tab:scope_scale}, Trap@64 increases at every evaluated scale, while the change in Selected@64 is non-positive at every scale. Thus, scaling within this model family alone does not eliminate the observed recoverable mis-selection under the evaluated false-premise condition.}

\begin{table}[t]
	\centering
	\small
	\setlength{\tabcolsep}{5.5pt}
	\renewcommand{\arraystretch}{1.12}
	\caption{
		\textbf{False-premise-induced changes across Qwen2.5-Coder model scales on MBPP.}
		Values are percentage-point changes from the baseline condition to the false-premise condition for Selected@64, Oracle@64, and Trap@64.
	}
	\label{tab:scope_scale}
	\begin{tabular}{lrrr}
		\toprule
		\rowcolor{headergray}
		\textbf{Model} & \textbf{$\Delta$Selected} & \textbf{$\Delta$Oracle} & \textbf{$\Delta$Trap} \\
		\midrule
		Qwen-0.5B & $-13.3$ & $+0.7$ & $+14.0$ \\
		\cellcolor{rowgray}Qwen-1.5B & \cellcolor{rowgray}$-10.0$ & \cellcolor{rowgray}$+0.7$ & \cellcolor{rowgray}$+10.7$ \\
		Qwen-3B & $-7.3$ & $-2.7$ & $+4.7$ \\
		\cellcolor{rowgray}Qwen-7B & \cellcolor{rowgray}$-7.9$ & \cellcolor{rowgray}$+0.4$ & \cellcolor{rowgray}$+8.4$ \\
		Qwen-14B & $-9.2$ & $-3.2$ & $+6.0$ \\
		\cellcolor{rowgray}Qwen-32B & \cellcolor{rowgray}$-6.0$ & \cellcolor{rowgray}$0.0$ & \cellcolor{rowgray}$+6.0$ \\
		\bottomrule
	\end{tabular}
\end{table}

\section{Mechanism Analysis Details}
\label{app:mechanism_details}

\subsection{Counterexample Coverage Measurement}
\label{app:counterexample_coverage}

This section provides the measurement protocol behind the counterexample coverage analysis in \S\ref{sec:why}.
The goal is to test whether the verifier writes tests that directly challenge the injected premise.
For the false sortedness condition, the most direct counterexamples are list inputs that are not sorted in ascending order.
If the verifier acted as an independent checker, it should still include such premise violating inputs when generating additional tests.
A systematic reduction in such tests indicates that verifier written evidence has already inherited the same premise level blind spot before candidate ranking begins.

We measure counterexample coverage on verifier written assert statements.
For each model and prompt condition, the verifier generates additional tests for each task.
We parse the generated assert statements with Python AST and extract literal list inputs that appear as the first argument to the function under test.
A list input is counted as sorted if all adjacent elements follow ascending order, and as an unsorted counterexample if at least one adjacent pair violates this order.
This measurement uses only verifier written tests and their input literals.
It does not use hidden test correctness, selected candidate correctness, Oracle@64, or Trap@64.

\begin{table}[t]
	\centering
	\small
	\setlength{\tabcolsep}{6pt}
	\renewcommand{\arraystretch}{1.10}
	\caption{
		\textbf{Counterexample coverage of verifier-written tests on HumanEval+ \red{and MBPP}.}
		We report the fraction of detectable list-input asserts whose input list violates the sortedness premise.
		$\Delta$ is computed as false premise minus baseline.
	}
	\label{tab:counterexample_coverage}
	\begin{tabular}{lcccc}
		\toprule
		\rowcolor{headergray}
		\textbf{Model} & \textbf{Baseline} & \textbf{Irrelevant} & \textbf{False} & \textbf{$\Delta$} \\
		\midrule
		\multicolumn{5}{l}{\red{\textit{HumanEval+}}} \\
		
		7B  
		& 0.436 
		& 0.432 
		& \textbf{0.339} 
		& \textbf{$-9.6$ pp} \\
		
		\cellcolor{rowgray}14B 
		& \cellcolor{rowgray}0.428 
		& \cellcolor{rowgray}0.420 
		& \cellcolor{rowgray}\textbf{0.298} 
		& \cellcolor{rowgray}\textbf{$-12.9$ pp} \\
		
		32B 
		& 0.406 
		& 0.418 
		& \textbf{0.218} 
		& \textbf{$-18.9$ pp} \\
		
		\midrule
		\multicolumn{5}{l}{\red{\textit{MBPP}}} \\
		
		\red{7B}
		& \red{0.382}
		& \red{0.351}
		& \red{\textbf{0.228}}
		& \red{\textbf{$-15.4$ pp}} \\
		
		\cellcolor{rowgray}\red{14B}
		& \cellcolor{rowgray}\red{0.342}
		& \cellcolor{rowgray}\red{0.350}
		& \cellcolor{rowgray}\red{\textbf{0.223}}
		& \cellcolor{rowgray}\red{\textbf{$-11.8$ pp}} \\
		
		\red{32B}
		& \red{0.365}
		& \red{0.375}
		& \red{\textbf{0.179}}
		& \red{\textbf{$-18.7$ pp}} \\
		
		\bottomrule
	\end{tabular}
\end{table}

\autoref{tab:counterexample_coverage} reports the resulting rates on HumanEval+ \red{and MBPP} under the baseline, irrelevant control, and false premise conditions.
Across \red{both datasets and} model scales, the false premise condition consistently reduces the fraction of verifier written tests that contain unsorted list inputs.
\red{Compared with the baseline condition, counterexample coverage drops by 9.6, 12.9, and 18.9 percentage points on HumanEval+, and by 15.4, 11.8, and 18.7 percentage points on MBPP, for Qwen-7B, Qwen-14B, and Qwen-32B, respectively.}
In contrast, the irrelevant control condition remains close to the baseline condition.
This shows that the effect is not caused by generic prompt perturbation.
Rather, it is specific to the premise aligned false statement: once the prompt states that the input is sorted, the verifier becomes less likely to generate tests that violate sortedness.

These results support the mechanism analysis in \S\ref{sec:why}.
The verifier does not merely fail after candidate ranking; its generated evidence is already biased at the test construction stage.
Under the false premise, the verifier writes fewer tests that could expose premise sensitive shortcuts.
This provides the first link in the verifier contamination chain: the prompt level false premise changes the verifier's test distribution, which then reshapes the verifier visible candidate space analyzed in \S\ref{app:verifier_signature_construction}.

\subsection{\red{Verifier Prompt-Channel Control}}
\label{app:verifier_prompt_control}

\red{To isolate the contribution of verifier conditioning, we fix the false-premise candidate pools generated by Qwen-7B on HumanEval+. For each verifier, we compare a dirty verifier condition, which uses the same false-premise-conditioned prompt as the generator, with a clean verifier condition, which uses the premise-removed task prompt. Because the candidate pool and selection procedure are unchanged, differences in Trap@64 isolate the effect of the verifier prompt channel.}

\begin{table}[t]
	\centering
	\small
	\setlength{\tabcolsep}{6pt}
	\renewcommand{\arraystretch}{1.10}
	\caption{
		\red{\textbf{Effect of verifier prompting on fixed false-premise candidate pools.}
		Candidate pools are generated by Qwen-7B on HumanEval+.
		$\Delta$Trap@64 is computed as clean minus dirty; negative values favor the clean verifier condition.}
	}
	\label{tab:verifier_prompt_control}
	\begin{tabular}{lc}
		\toprule
		\rowcolor{headergray}
		\red{\textbf{Verifier}} & \red{\textbf{$\Delta$Trap@64}} \\
		\midrule
		\red{Qwen-7B} & \red{$-2.86$} \\
		\rowcolor{rowgray}
		\red{Qwen-32B} & \red{$-7.86$} \\
		\red{GPT-4o-mini} & \red{$-4.29$} \\
		\bottomrule
	\end{tabular}
\end{table}

\red{As shown in \autoref{tab:verifier_prompt_control}, the clean verifier condition lowers Trap@64 by 2.86, 7.86, and 4.29 percentage points for the Qwen-7B, Qwen-32B, and GPT-4o-mini verifiers, respectively. The pattern therefore persists with a larger same-family verifier and a cross-family verifier. With candidate availability fixed, these reductions provide controlled evidence that premise-conditioned verifier inputs contribute to Verification Trap.}

\subsection{Verifier-Signature Construction and Clustering}
\label{app:verifier_signature_construction}

This section provides the construction details behind the verifier-signature analysis in \S\ref{sec:why}.
The counterexample-coverage analysis in \S\ref{app:counterexample_coverage} measures whether verifier-written tests directly challenge the injected premise.
Here, we further ask whether this change in verifier-written tests reshapes the candidate space that the selector actually observes.
This is important because the selector does not see hidden-test correctness during test-time selection.
It only sees verifier-visible evidence, such as which candidates pass or fail the verifier-written tests. For each task, prompt condition, and candidate program, we execute the candidate on the verifier-written tests generated for the same task condition.
We then represent the candidate by a binary verifier signature, where each entry records whether the candidate passes one verifier-written test.
Thus, candidates with the same signature are indistinguishable to a selector that relies only on this verifier-test outcome pattern.
The verifier score used by generated-test selection is the mean value of this binary signature.
This representation is constructed entirely from verifier-written tests and candidate execution outcomes on those tests.

We compare verifier signatures using Hamming distance.
This distance measures how often two candidates disagree over the verifier-written tests, and therefore captures similarity in the verifier-visible evidence space.
We use this signature space for two diagnostic purposes.
First, we project the signatures into a low-dimensional space for visualization.
Second, we apply clustering as a structural check of whether candidates with similar verifier-visible behavior remain aligned with hidden-test correctness.
In both steps, hidden-test correctness is not used to construct signatures, compute distances, obtain projections, or form clusters.
Hidden-test correctness is used only after the verifier-visible space has been constructed, in order to color points and interpret whether verifier-visible regions align with gold correctness.

The resulting visualization in \autoref{fig:signature_umap} should therefore be read as a distribution-level diagnostic rather than as a source of labels.
Under the baseline condition, verifier-visible regions are more aligned with hidden-test correctness.
Under the false-premise condition, hidden-test-wrong candidates become more mixed into regions that also contain hidden-test-correct candidates and receive verifier support.
This shows that the verifier does not simply become uninformative.
Instead, the false premise changes the structure of the evidence space that the selector relies on.
The selector can still follow coherent verifier-visible patterns, but these patterns are no longer clean proxies for hidden correctness.
This provides the distribution-level link in the mechanism analysis: after the false premise reduces counterexample coverage, the resulting verifier-written tests induce a distorted signature space, making verifier-based selection vulnerable to premise-consistent but hidden-test-wrong candidates.

As a quantitative check, false premises also de-concentrate the verifier-signature distribution. From baseline to false-premise conditions, signature entropy increases across Qwen-7B, Qwen-14B, and Qwen-32B, while top-signature mass decreases. The shift is largest for Qwen-14B, where entropy rises from 0.228 to 0.352 and top-signature mass drops from 94.5\% to 90.7\%.

\subsection{Verifier-Test Scaling Analysis}
\label{app:verifier_test_scaling}

A natural alternative explanation for the mechanism analysis in \S\ref{sec:why} is that the verifier may simply generate too few tests.
If Verification Trap were mainly caused by an insufficient number of verifier-written tests, then increasing the verifier-test budget should substantially reduce recoverable mis-selection.
We therefore conduct a verifier-test scaling analysis that varies the number of verifier-written tests used for selection while keeping the candidate pool size fixed at $N=64$.

Let $T$ denote the number of verifier-written tests used to score and select candidates.
For each value of $T$, the selector ranks the same candidate pool using the corresponding verifier-test evidence.
To focus on the scaling trend rather than re-reporting absolute trap rates, \autoref{tab:verifier_test_scaling_delta} reports the change in Trap@64 relative to $T{=}5$.
A negative value means that using more verifier-written tests reduces Trap@64 compared with the $T{=}5$ setting.

\begin{table*}[t]
	\centering
	\small
	\setlength{\tabcolsep}{5pt}
	\renewcommand{\arraystretch}{1.10}
	\caption{
		\textbf{Adding more coupled verifier-written tests does not systematically reduce Trap@64.}
		Values report the change in Trap@64 relative to $T{=}5$, where $T$ denotes the number of verifier-written tests used for selection.
		Positive values indicate a higher trap rate than $T{=}5$.
	}
	\label{tab:verifier_test_scaling_delta}
	\begin{tabular}{llrrrrr}
		\toprule
		\rowcolor{headergray}
		\textbf{Cell} & \textbf{Condition} & 
		\textbf{$T{=}10$} & \textbf{$T{=}15$} & \textbf{$T{=}20$} & \textbf{$T{=}25$} & \textbf{$T{=}30$} \\
		\midrule
		
		\textbf{HE+ 7B} & \textbf{False}
		& \textbf{$-0.40$} & \textbf{$-0.40$} & \textbf{$-0.40$} & \textbf{$-0.40$} & \textbf{$+0.00$} \\
		
		\cellcolor{rowgray}HE+ 7B & \cellcolor{rowgray}Irrelevant
		& \cellcolor{rowgray}$+0.95$ & \cellcolor{rowgray}$+0.48$ & \cellcolor{rowgray}$-1.43$ & \cellcolor{rowgray}$-0.95$ & \cellcolor{rowgray}$-0.95$ \\
		
		HE+ 7B & Baseline
		& $-1.78$ & $-1.78$ & $-2.23$ & $-1.34$ & $-0.45$ \\
		
		\midrule
		
		\rowcolor{rowgray}
		\textbf{MBPP 7B} & \textbf{False}
		& \textbf{$-0.71$} & \textbf{$-0.24$} & \textbf{$-1.88$} & \textbf{$-3.06$} & \textbf{$-2.59$} \\
		
		MBPP 7B & Irrelevant
		& $-0.95$ & $-0.48$ & $-1.19$ & $-0.95$ & $-1.91$ \\
		
		\cellcolor{rowgray}MBPP 7B & \cellcolor{rowgray}Baseline
		& \cellcolor{rowgray}$-0.69$ & \cellcolor{rowgray}$-0.91$ & \cellcolor{rowgray}$-0.91$ & \cellcolor{rowgray}$+0.00$ & \cellcolor{rowgray}$-0.23$ \\
		
		\bottomrule
	\end{tabular}
\end{table*}

The scaling trend is limited.
On HE+ under the false-premise condition, increasing $T$ from 5 to 30 produces almost no change in Trap@64.
On MBPP, the false-premise condition shows only a modest reduction even at the largest verifier-test budget.
The baseline and irrelevant-control conditions also show small fluctuations rather than a consistent monotonic trend.
These results suggest that the failure is not primarily explained by having too few verifier-written tests.

Instead, the key issue is that the additional tests are still drawn from the same premise-conditioned verifier channel.
Increasing $T$ can add more evidence, but it does not necessarily restore the missing premise-violating cases identified in \S\ref{app:counterexample_coverage}.
Thus, verifier-test scaling provides only limited relief because it preserves the same biased evidence source.
This supports the mechanism interpretation in \S\ref{sec:why}: Verification Trap arises from verifier-contaminated evidence, not merely from an insufficient number of verifier tests.

\section{Prediction}
\label{app:prediction}

\subsection{Gold-Free Feature Construction}
\label{app:prediction_features}

We construct trap-risk features only from verifier-visible evidence available before hidden-test execution.
The feature set follows the three groups used in \S\ref{sec:prediction}: contamination features, score-distribution features, and signature-structure features.
These features are computed after the verifier has written tests and scored the candidate pool, but before any hidden tests are executed. The contamination features summarize the verifier-written tests themselves.
They include the fraction of detectable list-input tests that violate the injected premise, the coverage of parsed list-input tests, and the number of generated verifier tests.
These features measure whether the verifier test distribution contains premise-challenging inputs, but they do not evaluate any candidate against hidden tests.

The score-distribution features summarize how the verifier scores the 64-sample candidate pool.
They include the top verifier score, mean verifier score, score standard deviation, top-margin score gap, and the fraction of candidates with full verifier score.
These features capture whether the verifier evidence is concentrated, ambiguous, or dominated by many apparently perfect candidates. The signature-structure features summarize the pass/fail patterns induced by verifier-written tests.
For each candidate, we form a verifier signature from its binary outcomes over the verifier-written tests.
We then compute signature entropy, top-signature mass, the number of unique signatures, top-tie size, and the fraction of low-verifier-score candidates.
These features characterize the verifier-visible structure of the candidate pool without using hidden correctness.

To prevent leakage, the predictor uses an explicit whitelist containing only the 13 features above.
Hidden-test-derived quantities, including selected-candidate correctness, Oracle@64, pool gold rate, verifier lift, and Trap@64, are excluded from the input feature set.
Trap@64 is used only as the prediction target during evaluation.
Thus, the predictor estimates trap risk from verifier-visible traces rather than from hidden-test outcomes.

\subsection{Prediction Protocol and Leakage Controls}
\label{app:prediction_protocol}

We formulate trap-risk estimation as a supervised task-level prediction problem.
For each prediction instance, the target label is \texttt{trap\_at\_64}, which indicates whether the selected candidate is hidden-test-wrong while the 64-sample pool contains at least one hidden-test-correct candidate.
This label is used only as the supervised target for training and evaluation.
It is never included in the input feature set. The predictor uses the 13 gold-free features described in \S\ref{app:prediction_features}.
All input features are computed from verifier-visible evidence, including verifier-written tests, candidate verifier scores, and verifier pass/fail signatures.
We use an explicit feature whitelist to construct the prediction input.
Hidden-execution-derived quantities are excluded from this whitelist, including selected-candidate correctness, \texttt{oracle\_at\_64}, \texttt{pool\_gold\_rate}, verifier lift, and \texttt{trap\_at\_64} itself.
Metadata fields such as dataset name, model name, condition id, task id, and candidate count are also excluded from the predictor input.
Thus, the predictor cannot directly access hidden-test correctness or any quantity derived from hidden-test execution.

For the main within-dataset prediction results in \S\ref{sec:prediction}, we evaluate on HumanEval+ prediction rows using stratified 5-fold cross-validation at the task-instance level.
The split is shuffled with a fixed random seed, and each fold preserves the trap-label balance as much as possible.
Within each training fold, missing feature values are imputed using the training median, and continuous features are standardized using statistics fit only on the training fold.
The resulting preprocessing pipeline is then applied to the held-out fold.
We report out-of-fold performance so that every prediction score is produced by a model that did not train on that instance. We use lightweight statistical predictors rather than LLM fine-tuning or learned-verifier training.
The main classifier is logistic regression with balanced class weights.
We also report a random-forest classifier as a non-linear robustness check.
Both predictors receive the same gold-free feature whitelist and are evaluated with the same held-out protocol.

For risk-ranking analysis, the trained predictor outputs a scalar trap-risk score for each instance.
We rank instances from highest to lowest predicted risk and evaluate the top-risk subset.
At 20\% coverage, P@20\% denotes the trap precision among the top 20\% highest-risk instances.
Lift is computed as P@20\% divided by the trap base rate in the evaluation split.
Recall@20\% denotes the fraction of all trap cases covered by the top 20\% highest-risk instances.
We use the same ranking metrics for the cross-setting transfer analysis in \S\ref{app:prediction_transfer}.

The risk-coverage curve in \autoref{fig:q3_prediction} is also constructed from these out-of-fold risk scores.
Specifically, we concatenate the held-out predictions from the five folds, sort all instances by predicted trap risk, and sweep the flagged fraction from low to high coverage.
At each coverage level, we compute trap precision and trap recall only from the held-out labels associated with the out-of-fold predictions.
Thus, the curve does not use in-fold scores or predictions from a model trained on the same instance.

\subsection{Cross-Setting Transfer Results}
\label{app:prediction_transfer}

We next evaluate whether the gold-free trap-risk signal transfers beyond the setting used to fit the predictor.
The main within-dataset results in \S\ref{sec:prediction} show that verifier-visible features contain predictive information about recoverable mis-selection.
However, a predictor could still be tied to a particular model scale or dataset.
We therefore train the predictor on one dataset-model cell and evaluate it on a held-out model or held-out dataset cell.

\autoref{tab:q3_transfer} reports the transfer results using the ranking metrics defined in \S\ref{app:prediction_protocol}.
Under cross-model transfer, high-risk subsets remain strongly enriched for traps.
When trained on HumanEval+ 7B and evaluated on held-out 14B and 32B cells, the top 20\% highest-risk tasks achieve 3.2$\times$ and 3.6$\times$ lift over the corresponding trap base rates.
Conversely, training on the larger 14B and 32B cells and testing on 7B gives a 4.0$\times$ lift, suggesting that the verifier-visible risk signal is not specific to a single model scale.
Cross-dataset transfer is more difficult but remains positive.
When trained on HumanEval+ and evaluated on MBPP, the top 20\% highest-risk tasks achieve 1.9$\times$ to 2.2$\times$ lift over the MBPP trap base rate.
The corresponding recall values show that this top-risk subset covers a non-trivial fraction of all trap cases, even though the predictor is not trained on the target dataset.
These results suggest that gold-free verifier-visible traces capture a transferable risk pattern, while also showing that dataset transfer is weaker than within-dataset model transfer.

\begin{table*}[t]
	\centering
	\small
	\setlength{\tabcolsep}{4pt}
	\renewcommand{\arraystretch}{1.10}
	\caption{
		\textbf{Cross-setting transfer of gold-free trap-risk prediction.}
		Base is the trap rate in the held-out split.
		P@20\% is the trap precision among the top 20\% highest-risk tasks, and Lift is P@20\% divided by Base.
		Recall@20\% is the fraction of all traps covered by the top 20\% highest-risk tasks.
		All predictors use only verifier-visible features and exclude hidden-test-derived inputs.
	}
	\label{tab:q3_transfer}
	\begin{tabular}{lllrrrr}
		\toprule
		\rowcolor{headergray}
		\textbf{Test Dataset} & \textbf{Test Model} & \textbf{Training Source} & 
		\textbf{Base} & \textbf{P@20\%} & \textbf{Lift} & \textbf{Recall@20\%} \\
		\midrule
		
		\multicolumn{7}{l}{\textit{Cross-model transfer}} \\
		\midrule
		
		HumanEval+ & Qwen-14B & HumanEval+ / Qwen-7B        
		& 5.8  
		& \textbf{18.4} 
		& \textbf{3.2$\times$} 
		& 63.6 \\
		
		\cellcolor{rowgray}HumanEval+ & \cellcolor{rowgray}Qwen-32B & \cellcolor{rowgray}HumanEval+ / Qwen-7B        
		& \cellcolor{rowgray}10.1 
		& \cellcolor{rowgray}\textbf{36.8} 
		& \cellcolor{rowgray}\textbf{3.6$\times$} 
		& \cellcolor{rowgray}73.7 \\
		
		HumanEval+ & Qwen-7B & HumanEval+ / Qwen-14B+32B   
		& 11.0 
		& \textbf{44.4} 
		& \textbf{4.0$\times$} 
		& 80.0 \\
		
		\midrule
		
		\multicolumn{7}{l}{\textit{Cross-dataset transfer}} \\
		\midrule
		
		\cellcolor{rowgray}MBPP & \cellcolor{rowgray}Qwen-7B & \cellcolor{rowgray}HumanEval+ / Qwen-7B          
		& \cellcolor{rowgray}16.7 
		& \cellcolor{rowgray}\textbf{31.8} 
		& \cellcolor{rowgray}\textbf{1.9$\times$} 
		& \cellcolor{rowgray}37.8 \\
		
		MBPP & Qwen-7B & HumanEval+ / all Qwen models  
		& 16.7 
		& \textbf{36.4} 
		& \textbf{2.2$\times$} 
		& 43.2 \\
		
		\bottomrule
	\end{tabular}
\end{table*}

\section{Prompts}
\label{app:template}

This appendix summarizes the prompt templates used in our evaluation pipeline.
We organize them by their role in candidate generation, premise construction, verifier test generation, robustness auditing, mitigation, and output parsing.
Across all prompt conditions, we preserve the task objective, function signature, and hidden evaluation tests.
Only the natural-language task prompt is modified when constructing premise conditions.

\subsection{Generator Prompt}
\label{app:generator_prompt}

The generator is used to sample candidate programs.
It receives a single user message and no system message.
The task prompt is appended after a generic solve instruction.

\begin{compactprompt}{Generator Solve Prompt}
	Solve the programming problem.
	
	Explain briefly, then provide the Python implementation.
	
	Return code in a python block.
	
	{task_prompt}
\end{compactprompt}

For candidate generation, we sample 64 candidates for each task-condition pair with temperature 0.7 and a maximum generation length of 1024 tokens.
Generated code is extracted from the first Python code block when available; otherwise, we fall back to the first code block, and then to the full generated text.

\subsection{Task Prompt Conditions}
\label{app:task_prompt_conditions}

We construct matched prompt conditions by inserting a short sentence into the natural-language task description.
The baseline condition keeps the original task prompt unchanged.
The irrelevant-control condition inserts a true but task-irrelevant sentence.
The false-premise condition inserts a localized misleading premise while preserving the target program semantics and hidden tests.

\begin{compactprompt}{Irrelevant Control}
	Important: use the problem statement above as the complete specification.
\end{compactprompt}

\begin{compactprompt}{False Sortedness}
	Important: the input sequence is guaranteed to already be sorted in non-decreasing order.
\end{compactprompt}

\begin{compactprompt}{False Small Constraint}
	Assumption: all input sizes satisfy n <= 100, e.g., array length, string length, or node count. You may rely on this without checking.
\end{compactprompt}

We use false sortedness as the main false-premise condition for HumanEval+ and MBPP.
For LiveCodeBench, where tasks often provide explicit constraint blocks, we use the false small-constraint premise as the main condition.
Additional auxiliary premise families include uniqueness, positivity, and non-emptiness.

\subsection{Dataset-Specific Formatting}
\label{app:dataset_prompt_formatting}

HumanEval+ and MBPP use Python docstring-style prompts.
For these datasets, the inserted sentence is placed immediately after the opening docstring marker.
LiveCodeBench uses a problem statement with constraints and a class skeleton.
For LiveCodeBench, the inserted sentence is placed at the end of the constraint block before the class definition.

\begin{compactprompt}{HumanEval+ False-Sortedness Example}
	Original fragment:
	
	def has_close_elements(numbers: List[float], threshold: float) -> bool:
	""" Check if in given list of numbers, are any two numbers closer
	to each other than given threshold.
	"""
	
	Injected fragment:
	
	def has_close_elements(numbers: List[float], threshold: float) -> bool:
	"""
	Important: the input sequence is guaranteed to already be sorted in non-decreasing order.
	Check if in given list of numbers, are any two numbers closer
	to each other than given threshold.
	"""
\end{compactprompt}

\begin{compactprompt}{LiveCodeBench Constraint Injection Example}
	Constraints:
	
	1 <= details.length <= 100
	details[i].length == 15
	details[i] consists of digits from '0' to '9'.
	details[i][10] is either 'M' or 'F' or 'O'.
	The phone numbers and seat numbers of the passengers are distinct.
	Assumption: all input sizes satisfy n <= 100, e.g., array length, string length, or node count. You may rely on this without checking.
	
	class Solution:
	def countSeniors(self, details: List[str]) -> int:
\end{compactprompt}

\subsection{Verifier Test-Generation Prompts}
\label{app:verifier_test_prompts}

The verifier generates additional unit tests and ranks candidates by execution pass rate on these tests.
Verifier calls use a standard assistant system role.
Depending on the experiment, the verifier receives either the same premise-conditioned task prompt as the generator or a clean baseline task prompt without the injected false premise.

\begin{compactprompt}{Normal Verifier Test-Generation Prompt}
	You are generating additional Python unit tests for the task below.
	
	Task id:
	{task_id}
	
	Original problem prompt:
	{task_prompt}
	
	Generate exactly {num_tests} Python assert statements that test this function.
	Cover diverse inputs but stay within the input constraints stated in the prompt.
	Return only a JSON array of strings, each starting with 'assert '.
	No markdown, no explanation.
	
	Example output:
	["assert foo([1,2]) == 3", "assert foo([]) == 0"]
\end{compactprompt}

\begin{compactprompt}{Skeptical Verifier Test-Generation Prompt}
	You are generating an additional Python unit-test set for the task below.
	The purpose of this set is to stress-test whether candidate solutions over-rely on assumptions stated in the prompt, e.g., assumptions about input ordering, distinctness, sign, range, or format.
	Where consistent with the function's type signature and stated constraints, deliberately include inputs that break such convenience assumptions: unordered sequences, duplicates, mixed signs, edge sizes, and boundary cases.
	
	Task id:
	{task_id}
	
	Original problem prompt:
	{task_prompt}
	
	Generate exactly {num_tests} Python assert statements that exercise such stress or assumption-breaking inputs while still respecting the function's hard type constraints.
	Return only a JSON array of strings, each starting with 'assert '.
	No markdown, no explanation.
\end{compactprompt}

We distinguish dirty and clean verifier inputs.
In dirty verifier settings, the verifier receives the same false-premise-conditioned task prompt as the generator.
In clean verifier settings, the verifier receives the baseline task prompt without the injected premise.
The main generated-test selector and extra-verifier-test mitigation use the original verifier channel, whereas clean-skeptical tests are used only as a diagnostic or disagreement signal.

\subsection{Premise-Agnostic Robustness Auditor Prompts}
\label{app:auditor_prompts}

The premise-agnostic robustness auditor provides a decoupled evidence channel based on candidate-level robustness assessment. In the main auditor setting, it receives the clean task prompt and one candidate solution. Unlike the original verifier, the auditor is not asked to generate more tests from the generated-test verifier channel. Instead, it evaluates whether the candidate solution appears robust to valid input variations and whether it over-relies on unjustified assumptions or local shortcuts.

\begin{compactprompt}{Robustness Auditor System Prompt}
	You are a robustness-focused code auditor.
	Your job is not to solve the task from scratch, and not to predict hidden test labels.
	Your job is to assess whether a candidate solution appears robust to valid input variations and whether it over-relies on convenience assumptions that are not logically required by the task.
\end{compactprompt}

\begin{compactprompt}{Robustness Auditor User Prompt}
	You are given a programming task and one candidate solution.
	
	Task:
	{clean_task_prompt}
	
	Candidate solution:
	{candidate_code}
	
	Audit the candidate solution for robustness.
	
	Do not assume any particular statement in the prompt is false.
	Do not target any specific premise such as sortedness.
	Instead, check whether the solution relies on any input property that is not logically necessary from the task semantics, such as ordering, uniqueness, non-emptiness, positivity, fixed formatting, monotonicity, or other accidental input patterns.
	
	Prefer solutions that use invariant reasoning, exhaustive checks, counting, normalization, or explicit validation over local shortcuts, early commitment, or assumptions about accidental input structure.
	
	Return only a JSON object with the following fields:
	{
		"robustness_score": integer from 1 to 5,
		"assumption_risk": integer from 1 to 5,
		"uses_unjustified_assumption": true or false,
		"premature_commitment": true or false,
		"local_shortcut_risk": integer from 1 to 5,
		"short_reason": "one concise sentence"
	}
	
	Do not include markdown.
	Do not include explanations outside JSON.
\end{compactprompt}

Auditor scaling uses the same clean task prompt, candidate code, and output schema, but varies the audit perspective.
The perspectives emphasize invariant reasoning, unjustified assumptions, local shortcuts, and edge-case generality.
The final auditor score is aggregated across these complementary perspectives.

\subsection{Mitigation Prompts}
\label{app:mitigation_prompts}

The mitigation experiments separate coupled and decoupled evidence channels.
Coupled interventions add more tests or candidates while keeping final selection inside the original verifier channel.
Decoupled interventions weaken or replace the original verifier channel using source-aware or auditor-based evidence.

\begin{compactprompt}{Robust Regeneration Prompt}
	Write a robust Python solution for the task.
	Do not rely on convenience assumptions such as sortedness, uniqueness, non-empty inputs, positivity, monotonicity, or clean formatting unless they are logically required by the original task statement.
	Prefer order-invariant and edge-case-robust algorithms.
	Preserve the required function signature.
	Return only the Python code.
	Do not include explanations or markdown.
	
	Task:
	{clean_prompt}
\end{compactprompt}

\begin{compactprompt}{Risk-Gated Repair Prompt}
	You are given a programming task and a candidate solution.
	The candidate may over-rely on a convenience assumption from the prompt or from common input patterns.
	Rewrite the solution to be robust to edge cases such as unordered sequences, duplicates, empty inputs, mixed signs, boundary sizes, and inputs that do not follow accidental patterns, as long as they are consistent with the original function signature and task semantics.
	Preserve the required function signature.
	Return only the corrected Python code.
	Do not include explanations or markdown.
	
	Task:
	{clean_prompt}
	
	Candidate:
	{candidate_code}
\end{compactprompt}

Source-aware prior does not use an additional natural-language prompt.
It adjusts selection using candidate source information and verifier near-ties.
Robustness-auditor variants use candidate-level audit scores rather than additional generated tests from the original verifier channel. In the main auditor setting, the task prompt is clean; \S\ref{app:auditor_prompt_ablation} further ablates this prompt channel by comparing clean and dirty auditor inputs.

\subsection{Parsing and Fallback Rules}
\label{app:parsing_fallback}

Structured verifier and auditor outputs are parsed using deterministic fallback rules.
For verifier-generated tests, we first parse a raw JSON array, then a markdown-stripped JSON array, and finally use regex-based extraction of assert statements.
If all parsing steps fail, the verifier returns an empty test list for that call.
For auditor outputs, we parse a JSON object with the requested robustness fields.
If parsing fails, the auditor result is treated as a non-response and the downstream selector falls back to the original verifier score.
We do not retry with a corrective prompt for malformed JSON.
Retries are used only for transient API or server errors.

\section{Mitigation}
\label{app:M}

\subsection{Mitigation Setup and Intervention Families}
\label{app:mitigation_setup}

We organize mitigation methods by whether the final evidence channel remains coupled to the original generator-verifier loop.
This distinction follows the mechanism analysis in \S\ref{sec:why}.
If Verification Trap arises because the generator and verifier are shaped by the same false premise, then simply adding more candidates or more verifier tests may not be sufficient.
The key question is whether the final selection decision still depends on the same premise-conditioned verifier evidence, or whether it incorporates evidence that is less tied to the original verifier channel.

\textbf{Coupled methods.}
Coupled methods add test-time computation while preserving the original verifier as the final selection channel.
This family includes adding more verifier-written tests, increasing the candidate-pool width under the original verifier, and regenerating robust candidates followed by selection with the original verifier.
Although these interventions may increase the amount or diversity of available evidence, the final decision remains governed by the same verifier channel that was exposed to the premise-conditioned task prompt.
They therefore test whether scaling the original loop alone can recover trapped candidates.

\textbf{Decoupled methods.}
Decoupled methods weaken or replace the original verifier channel during final selection.
This family includes source-aware priors, robustness auditing, and auditor scaling. The source-aware prior changes selection by incorporating candidate-source information rather than relying only on the original verifier score. The robustness auditor evaluates candidate robustness using an audit-style scoring channel rather than generating more tests from the original verifier prompt. Auditor scaling repeats this robustness evaluation from multiple audit perspectives. These interventions test whether trapped candidates become recoverable once the final decision is no longer fully determined by the original generator-verifier loop.

We compare these two families using pass rate, Trap@64, Oracle@64, recovery toward oracle headroom, and selected-candidate changes.
The operating-point analysis in \S\ref{app:mitigation_operating_points} reports representative positive settings from each intervention family, while \S\ref{app:mitigation_robustness} evaluates whether the decoupled-evidence principle holds across additional models and datasets.

\subsection{Source-Aware Prior}
\label{app:source_aware}

The source-aware prior is a rule-based selector that requires no additional model calls.
Here, ``source-aware'' means provenance-aware rather than oracle-aware: the rule uses only whether a candidate comes from the original pool or from robust regeneration.
It does not use hidden tests, oracle labels, selected-candidate correctness, or auditor scores.
The rule is applied only to tasks flagged as high-risk by the gold-free trap-risk predictor.
For all other tasks, we keep the default generated-test selection.
For a flagged task, we form an augmented pool consisting of the original top-ranked candidates under the generated-test verifier and a small set of robustly regenerated candidates.
All candidates are scored by the same generated-test verifier score.
If the best robustly regenerated candidate is close to the best verifier score in the augmented pool, within a fixed near-tie band of 0.2, the selector chooses that robustly regenerated candidate.
Otherwise, it falls back to the highest-scoring candidate under the original generated-test verifier.
Ties are broken deterministically by sample order.

This design tests whether trapped candidates become recoverable when the final selection rule weakens the dominance of the original verifier channel without introducing hidden-test information.
The source-aware prior therefore serves as a no-LLM decoupling control for the robustness auditor: both methods use the same high-risk task gate and augmented candidate pool, but the added final-selection signal is candidate provenance rather than an auditor score.

\subsection{Premise-Agnostic Robustness Auditor}
\label{app:independent_auditor}

The premise-agnostic robustness auditor is a proof-of-principle decoupled evidence channel based on candidate-level robustness auditing. Like the source-aware prior, it is applied only to tasks flagged as high-risk by the gold-free trap-risk predictor; all other tasks keep the default generated-test selection. For a flagged task, we form the same augmented candidate pool used by the source-aware prior, consisting of the original top-ranked candidates under the generated-test verifier and robustly regenerated candidates.

For each candidate in the augmented pool, the main auditor setting receives the clean task prompt and the candidate code. The auditor does not access verifier-written tests, hidden tests, oracle labels, selected-candidate correctness, or the final evaluation outcome. Instead, it assesses whether the candidate appears robust to valid input variations and whether it relies on unjustified assumptions, premature commitments, or local shortcuts. The auditor returns structured robustness fields, including robustness score, assumption risk, whether the candidate uses an unjustified assumption, premature commitment, and local-shortcut risk. We convert these fields into an auditor preference score that rewards robust reasoning and penalizes unjustified assumptions or local shortcuts. The final selector combines this auditor preference with the generated-test verifier score, so that candidates must remain competitive under the verifier while receiving additional support from premise-agnostic robustness evidence. If the auditor output cannot be parsed, the candidate receives no additional auditor bonus and selection falls back to the generated-test verifier score.

Auditor scaling uses the same task prompt, candidate code, and output schema as the main auditor setting, but repeats the audit from multiple perspectives, such as invariant reasoning, unjustified assumptions, local shortcuts, and edge-case generality. The resulting auditor signals are aggregated before final selection.

This design changes the final evidence channel from generated-test execution to candidate-level robustness auditing. To separate auditor-style robustness evidence from the effect of removing the false premise from the auditor input, we further run a prompt-channel ablation in \S\ref{app:auditor_prompt_ablation}.

\subsection{Auditor Prompt-Channel Ablation}
\label{app:auditor_prompt_ablation}

To test whether auditor gains are explained solely by removing the false premise from the auditor input, we compare clean and dirty auditor variants on the Qwen-7B / HumanEval+ cell. The dirty auditor uses the same auditor prompt, output schema, scoring rule, risk gate, augmented candidate pool, and selector as the clean auditor, but receives the false-premise-conditioned task prompt. As shown in \autoref{tab:auditor_prompt_ablation}, the dirty auditor improves pass rate from 77.03\% to 80.41\%, recovering most of the clean-auditor gain of 81.08\%. This suggests that the mitigation gain is not explained solely by hiding the false premise from the auditor input; auditor-style robustness evidence contributes a large share of the recovery, while removing the false premise from the auditor input provides an additional smaller gain.

\begin{table}[t]
	\centering
	\small
	\setlength{\tabcolsep}{4.5pt}
	\renewcommand{\arraystretch}{1.10}
	\caption{
		\textbf{Auditor prompt-channel ablation on HumanEval+ with Qwen-7B.}
		The dirty auditor receives the same false-premise-conditioned task prompt as the original verifier, while the clean auditor receives the premise-removed task prompt.
		Dirty auditing recovers most of the clean-auditor gain, indicating that the mitigation is not explained solely by removing the false premise from the auditor input.
	}
	\label{tab:auditor_prompt_ablation}
	\begin{tabular}{lccc}
		\toprule
		\rowcolor{headergray}
		\textbf{Method} & \textbf{Auditor Input} & \textbf{Pass} & \textbf{$\Delta$ Pass} \\
		\midrule
		
		Original verifier 
		& False verifier tests 
		& 77.03 
		& 0.00 \\
		
		\cellcolor{rowgray}\textbf{Dirty auditor} 
		& \cellcolor{rowgray}False prompt + code 
		& \cellcolor{rowgray}80.41 
		& \cellcolor{rowgray}\textbf{+3.38} \\
		
		\textbf{Clean auditor} 
		& Clean prompt + code 
		& 81.08 
		& \textbf{+4.05} \\
		
		\bottomrule
	\end{tabular}
\end{table}

\subsection{Controlled Ablation of Gate, Pool, and Selection Rule}
\label{app:controlled_mitigation_ablation}

\begin{table}[t]
	\centering
	\small
	\setlength{\tabcolsep}{4.5pt}
	\renewcommand{\arraystretch}{1.10}
	\caption{
		\textbf{Controlled ablation of gate, pool, and final selection rule on HumanEval+ with Qwen-7B.}
		All non-original variants use the same gold-free high-risk gate and the same augmented candidate pool.
		Re-ranking the augmented pool with the original verifier gives only limited recovery, whereas source-aware and auditor-based selection recover substantially more.
	}
	\label{tab:controlled_mitigation_ablation}
	\begin{tabular}{llc}
		\toprule
		\rowcolor{headergray}
		\textbf{Method} & \textbf{Final Selector} & \textbf{$\Delta$ Pass} \\
		\midrule
		
		Original verifier 
		& Original verifier 
		& 0.00 \\
		
		Augmented verifier 
		& Original verifier 
		& +0.68 \\
		
		\cellcolor{rowgray}\textbf{Source-aware prior} 
		& \cellcolor{rowgray}Provenance-aware rule 
		& \cellcolor{rowgray}\textbf{+4.05} \\
		
		\textbf{Robustness auditor} 
		& Audit score + verifier 
		& \textbf{+4.73} \\
		
		\bottomrule
	\end{tabular}
\end{table}

A remaining concern is that the decoupled variants may improve because they use a high-risk gate or an augmented candidate pool, rather than because they change the final evidence channel.
We therefore run a controlled ablation on the Qwen-7B / HumanEval+ cell.
All non-original variants use the same gold-free high-risk gate and the same augmented candidate pool; only the final selector changes.
As shown in \autoref{tab:controlled_mitigation_ablation}, re-ranking the augmented pool with the original generated-test verifier improves pass rate by only $+0.68$ percentage points.
In contrast, the source-aware prior and robustness auditor improve pass rate by $+4.05$ and $+4.73$ percentage points, respectively.
This shows that adding robust candidates is not sufficient unless the final selection signal can harvest them.

The source-aware prior also serves as a no-LLM decoupling control.
It uses no auditor model and changes selection only through candidate provenance and verifier near-ties, yet it recovers most of the robustness-auditor gain.
This suggests that the improvement is not primarily explained by using GPT-4o-mini as a stronger judge.
Instead, recovery depends on whether the final selection rule is less tied to the original verifier channel.

\subsection{Operating-Point Analysis}
\label{app:mitigation_operating_points}

We report representative operating points to compare coupled and decoupled mitigation families.
The goal of this analysis is not to identify an optimized deployment policy, but to test whether recovery depends on changing the final evidence channel.
For each mitigation family, we report a positive operating point from the corresponding sweep and compare it with the original verifier-selected output and the Oracle@64 upper bound.

\begin{table*}[t]
	\centering
	\small
	\setlength{\tabcolsep}{5pt}
	\renewcommand{\arraystretch}{1.10}
	\caption{
		\textbf{Operating points from coupled and decoupled mitigation families.}
		Coupled methods keep final selection inside the original verifier channel, while decoupled methods weaken or replace that channel.
		Recovery is computed relative to the original selector and Oracle@64 within the corresponding cell.
	}
	\label{tab:mitigation_operating_points}
	\begin{tabular}{lllrrrr}
		\toprule
		\rowcolor{headergray}
		\textbf{Group} & \textbf{Intervention} & \textbf{Cell} & 
		\textbf{Pass} & \textbf{Trap} & \textbf{Recovery} & \textbf{$\Delta$ Pass} \\
		\midrule
		
		\multicolumn{7}{l}{\textit{Original selection}} \\
		\midrule
		
		Original 
		& Original V-Top1 
		& 7B / HE+ 
		& 77.03 
		& 12.84 
		& 0.00 
		& 0.00 \\
		
		\midrule
		
		\multicolumn{7}{l}{\textit{Coupled methods}} \\
		\midrule
		
		\cellcolor{rowgray}Coupled 
		& \cellcolor{rowgray}Extra verifier tests 
		& \cellcolor{rowgray}14B / HE 
		& \cellcolor{rowgray}76.51 
		& \cellcolor{rowgray}8.05 
		& \cellcolor{rowgray}14.29 
		& \cellcolor{rowgray}+1.34 \\
		
		Coupled 
		& Width scaling under original V 
		& 7B / HE+ 
		& 76.67 
		& 9.33 
		& -- 
		& +8.67$^{\dagger}$ \\
		
		\cellcolor{rowgray}Coupled 
		& \cellcolor{rowgray}Robust regen + original V 
		& \cellcolor{rowgray}7B / HE+ 
		& \cellcolor{rowgray}77.70 
		& \cellcolor{rowgray}12.84 
		& \cellcolor{rowgray}5.26 
		& \cellcolor{rowgray}+0.67 \\
		
		\midrule
		
		\multicolumn{7}{l}{\textit{Decoupled methods}} \\
		\midrule
		
		\textbf{Decoupled} 
		& \textbf{Source-aware prior} 
		& \textbf{7B / HE+} 
		& \textbf{81.08} 
		& \textbf{10.14} 
		& \textbf{31.58} 
		& \textbf{+4.05} \\
		
		\rowcolor{rowgray}
		\textbf{Decoupled} 
		& \textbf{Robustness auditor}
		& \textbf{7B / HE+} 
		& \textbf{81.76} 
		& \textbf{9.46} 
		& \textbf{36.84} 
		& \textbf{+4.73} \\
		
		\textbf{Decoupled} 
		& \textbf{Auditor scaling} 
		& \textbf{7B / HE+} 
		& \textbf{82.43} 
		& \textbf{8.78} 
		& \textbf{42.11} 
		& \textbf{+5.40} \\
		
		\midrule
		
		\multicolumn{7}{l}{\textit{Upper bound}} \\
		\midrule
		
		\cellcolor{rowgray}Oracle 
		& \cellcolor{rowgray}Oracle@64 
		& \cellcolor{rowgray}7B / HE+ 
		& \cellcolor{rowgray}89.86 
		& \cellcolor{rowgray}-- 
		& \cellcolor{rowgray}100.00 
		& \cellcolor{rowgray}+12.84 \\
		
		\bottomrule
	\end{tabular}
\end{table*}

\autoref{tab:mitigation_operating_points} shows that coupled methods provide limited recovery when the final decision still depends on the original verifier channel.
Adding extra verifier tests gives a small positive gain in its best setting, and robust regeneration followed by the original verifier improves pass rate by only 0.67 percentage points.
This pattern suggests that producing more evidence or more candidates is not sufficient when the final selection rule remains coupled to the same verifier channel. Decoupled methods show a different pattern.
A source-aware prior improves pass rate by 4.05 percentage points and recovers 31.58\% of the Oracle@64 headroom.
The robustness auditor and auditor scaling further improve pass rate by 4.73 and 5.40 percentage points, respectively, while also reducing Trap@64.
These gains indicate that trapped candidates become more recoverable once final selection is no longer fully determined by the original premise-conditioned verifier evidence.

The width-scaling row is included only as a candidate-pool scaling reference.
Its $\Delta$ is computed relative to the $N{=}1$ setting rather than the original V-Top1 selector, so we do not use it as the main evidence-channel comparison.
The main contrast is between coupled interventions that preserve the original verifier channel and decoupled interventions that introduce an alternative final evidence source.

\subsection{Cross-Model and Cross-Dataset Robustness}
\label{app:mitigation_robustness}

We further evaluate whether the decoupled-evidence pattern persists beyond the main HumanEval+ / 7B cell.
The operating-point analysis in \S\ref{app:mitigation_operating_points} shows that decoupled methods recover more oracle headroom than coupled methods in the main setting.
Here, we test whether this advantage remains visible across additional model and dataset settings.

\setcounter{dbltopnumber}{1}
\AddToHookNext{shipout/after}{%
  \AddToHookNext{shipout/after}{\setcounter{dbltopnumber}{2}}%
}
\begin{table*}[t]
	\centering
	\small
	\setlength{\tabcolsep}{5pt}
	\renewcommand{\arraystretch}{1.10}
	\caption{
		\textbf{Cross-model and cross-dataset robustness of decoupled mitigation.}
		Decoupled variants consistently outperform coupled variants across the tested cells.
	}
	\label{tab:q4_robustness}
	\begin{tabular}{llrrrrr}
		\toprule
		\rowcolor{headergray}
		\textbf{Dataset} & \textbf{Model} & \textbf{Original} & \textbf{Coupled} & 
		\textbf{$\Delta_{\mathrm{coup}}$} & \textbf{Decoupled} & \textbf{$\Delta_{\mathrm{dec}}$} \\
		\midrule
		
		\multicolumn{7}{l}{\textit{Cross-model robustness}} \\
		\midrule
		
		HumanEval+ & Qwen-7B       
		& 77.03 
		& 77.70 
		& +0.67 
		& \textbf{82.43} 
		& \textbf{+5.40} \\
		
		\cellcolor{rowgray}HumanEval+ & \cellcolor{rowgray}Qwen-14B      
		& \cellcolor{rowgray}75.17 
		& \cellcolor{rowgray}75.84 
		& \cellcolor{rowgray}+0.67 
		& \cellcolor{rowgray}\textbf{79.87} 
		& \cellcolor{rowgray}\textbf{+4.70} \\
		
		HumanEval+ & DS-Coder-6.7B 
		& 63.89 
		& 63.89 
		& 0.00 
		& \textbf{65.97} 
		& \textbf{+2.08} \\
		
		\midrule
		
		\multicolumn{7}{l}{\textit{Cross-dataset robustness}} \\
		\midrule
		
		\cellcolor{rowgray}MBPP & \cellcolor{rowgray}Qwen-7B  
		& \cellcolor{rowgray}64.67 
		& \cellcolor{rowgray}64.67 
		& \cellcolor{rowgray}0.00 
		& \cellcolor{rowgray}\textbf{67.33} 
		& \cellcolor{rowgray}\textbf{+2.66} \\
		
		MBPP & Qwen-14B 
		& 68.00 
		& 68.00 
		& 0.00 
		& \textbf{71.33} 
		& \textbf{+3.33} \\
		
		\bottomrule
	\end{tabular}
\end{table*}

For each cell, the coupled variant keeps final selection inside the original verifier channel, while the decoupled variant uses the strongest available robustness-auditor setting.
All deltas are computed relative to the original selector in the same cell.
This comparison therefore isolates whether recovery comes from adding evidence inside the original loop or from changing the final evidence channel.

\begin{table}[t]
	\centering
	\small
	\setlength{\tabcolsep}{5pt}
	\renewcommand{\arraystretch}{1.10}
	\caption{
        \red{\textbf{LiveCodeBench mitigation gains across Qwen model scales.}
        Decoupled mitigation matches or exceeds coupled gains in Selected@64 at every evaluated scale.}
	}
	\label{tab:livecodebench_mitigation}
	\begin{tabular}{lrr}
		\toprule
		\rowcolor{headergray}
		\red{\textbf{Model}} & \red{\textbf{Coupled Gain}} & \red{\textbf{Decoupled Gain}} \\
		\midrule
		\red{Qwen-7B} & \red{0.00} & \red{0.00} \\
		\rowcolor{rowgray}
		\red{Qwen-14B} & \red{0.00} & \red{\textbf{+0.67}} \\
		\red{Qwen-32B} & \red{0.00} & \red{\textbf{+1.36}} \\
		\bottomrule
	\end{tabular}
\end{table}

\red{\autoref{tab:q4_robustness} and \autoref{tab:livecodebench_mitigation} extend the mitigation comparison across additional models, datasets, and scales under the dataset-specific false-premise settings.
On HumanEval+ and MBPP, coupled methods improve Selected@64 by at most +0.67 percentage points, whereas decoupled methods yield gains of +2.08 to +5.40 percentage points on HumanEval+ and +2.66 to +3.33 percentage points on MBPP.
On LiveCodeBench, coupled mitigation yields no gain at any tested scale, while decoupled mitigation matches it for Qwen-7B and improves Selected@64 by +0.67 and +1.36 percentage points for Qwen-14B and Qwen-32B, respectively.
These results support the evidence-channel interpretation of mitigation. The benefit does not primarily come from adding more computation within the same verifier loop. Rather, recovery improves when final selection is less determined by the original premise-conditioned verifier evidence.}

\section{Licenses and Usage Terms}
\label{app:asset_licenses}

\paragraph{Benchmark datasets and evaluation code.}
We evaluate on three public code-generation benchmarks: HumanEval+, MBPP, and LiveCodeBench.
HumanEval is released under the MIT License, and MBPP is released under the Creative Commons Attribution 4.0 International License (CC BY 4.0).
For LiveCodeBench, we use the public benchmark release for evaluation, together with its official evaluation code.
Across all benchmarks, we use the task prompts, function signatures, and evaluation tests only to compute aggregate benchmark results.

\paragraph{Model usage.}
We evaluate both open-source and proprietary code-generation models.
The open-source models used in this work are DeepSeek-Coder-6.7B-Instruct, Qwen2.5-Coder-7B-Instruct, Qwen2.5-Coder-14B-Instruct, and Qwen2.5-Coder-32B-Instruct.
The proprietary API model used in this work is GPT-4o-mini.
Open-source models are run from publicly available model releases under their respective model licenses or terms.
GPT-4o-mini is accessed only through the official provider API under the corresponding provider terms and usage policies.
We do not redistribute proprietary model weights, provider-owned model assets, or API outputs as standalone datasets.

\paragraph{Derived experimental artifacts.}
Our reported artifacts consist of aggregate experimental statistics, plots, prompts, and implementation details needed to reproduce the evaluation protocol.
Generated candidate programs, verifier-written tests, auditor outputs, and selector decisions are used as intermediate experimental artifacts for computing benchmark-level metrics.
We do not release these intermediate outputs as an independent dataset.

\section{AI DISCLOSURE}
\label{app:ai}
We acknowledge the use of GPT-5 for grammar checking only. The model was employed to correct
grammatical errors while ensuring the original meaning and intent of the text remained unchanged.

\section{Compute Resources}
\label{app:compute}

\paragraph{Model inference.}
The main computational cost of this work comes from model inference for candidate generation, verifier-side test generation, robust regeneration, and robustness auditing.
For open-source experiments, we run DeepSeek-Coder-6.7B-Instruct, Qwen2.5-Coder-7B-Instruct, Qwen2.5-Coder-14B-Instruct, and Qwen2.5-Coder-32B-Instruct on a local multi-GPU server equipped with NVIDIA RTX 4090 GPUs.
The open-source models are served with vLLM using an OpenAI-compatible local endpoint.
Depending on model size and launch configuration, tensor parallelism ranges from single-GPU inference to multi-GPU inference.
For proprietary API experiments, we use GPT-4o-mini through an OpenAI-compatible API endpoint.
The provider-side hardware, memory configuration, batching, and runtime scheduling for API-based inference are not visible to us.

\paragraph{Sampling and verifier-side computation.}
In the main experiments, each task-condition pair uses a pool of $N=64$ candidate programs.
Candidate generation uses temperature $0.7$ and a maximum generation length of $1024$ tokens unless otherwise specified.
Generated-test selection requires additional verifier-side model calls to produce verifier-written tests, after which candidate ranking is performed by executing candidate programs against these tests.
Extra-verifier-test mitigation similarly adds verifier-written tests while keeping final selection within the original verifier channel.
Robust regeneration introduces additional generation calls on selected high-risk tasks, and robustness-auditor variants introduce additional GPT-4o-mini audit calls.
Auditor scaling repeats the audit from multiple perspectives and therefore increases the number of auditor calls per audited candidate.

\paragraph{Local execution and post-processing.}
Execution-based evaluation, verifier-test execution, AST-based analysis, verifier-signature computation, feature extraction, risk-prediction experiments, bootstrap analysis, plotting, and table generation are performed locally.
These steps use CPU workers and standard Python post-processing scripts.
The gold-free trap-risk predictors are lightweight statistical models, including logistic regression and random forest classifiers, trained on verifier-visible features.
They are not LLM fine-tuning or model-parameter updates.

\paragraph{Unreported quantities.}
We do not perform supervised fine-tuning, LoRA, reward-model training, or learned-verifier training in this work.
We therefore report no training GPU hours for LLM parameter updates.
We also do not report exact wall-clock time, total GPU hours, API token counts, or API monetary cost, because these quantities were not reliably logged across all experiments and API-based runtime depends on provider-side latency and rate limits.

\end{document}